\documentclass[Afour,sagev,times]{sagej}
\DeclareUnicodeCharacter{2093}{$_x$}

\makeatletter
\NAT@superfalse
\makeatother
\setcitestyle{numbers,square,comma,sort&compress}

\usepackage{moreverb,url}
\usepackage[colorlinks,bookmarksopen,bookmarksnumbered,citecolor=red,urlcolor=red]{hyperref}
\newcommand\BibTeX{{\rmfamily B\kern-.05em \textsc{i\kern-.025em b}\kern-.08em
T\kern-.1667em\lower.7ex\hbox{E}\kern-.125emX}}
\usepackage{color}
\usepackage{graphicx}
\usepackage{caption}
\usepackage{mathtools}
\usepackage{url}
\usepackage{graphicx} 
\usepackage{float}
\usepackage{caption}
\usepackage{subcaption}
\usepackage{textcomp}
\usepackage{xcolor}
\usepackage{amssymb,amsfonts}
\usepackage{amsmath}
\usepackage{tikz}
\usetikzlibrary{shapes.geometric, arrows, positioning}
\usetikzlibrary{arrows.meta, positioning, shadows.blur}
\usepackage{multirow}
\usepackage{algorithm}
\usepackage{algpseudocode}

\begin{document}

\bibliographystyle{SageV} 

\runninghead{Zinage, et al.}

\title{Bayesian Calibration of Engine-out NOx Models for Engine-to-Engine Transferability}

\author{Shrenik Zinage\affilnum{1}, Peter Meckl\affilnum{1}, Ilias Bilionis\affilnum{1}}

\affiliation{\affilnum{1}School of Mechanical Engineering, Purdue University, West Lafayette, IN, USA}

\corrauth{Ilias Bilionis}

\email{ibilion@purdue.edu}

\begin{abstract}
Accurate prediction of engine-out NOx is essential for meeting stringent emissions regulations and optimizing engine performance. Traditional approaches rely on models trained on data from a small number of engines, which can be insufficient in generalizing across an entire population of engines due to sensor biases and variations in input conditions. In real world applications, these models require tuning or calibration to maintain acceptable error tolerance when applied to other engines. This highlights the need for models that can adapt with minimal adjustments to accommodate engine-to-engine variability and sensor discrepancies. 
While previous studies have explored machine learning methods for predicting engine-out NOx, these approaches often fail to generalize reliably across different engines and operating environments.
To address these issues, we propose a Bayesian calibration framework that combines Gaussian processes (GP) with approximate Bayesian computation to infer and correct sensor biases. Starting with a pre-trained model developed using nominal engine data, our method identifies engine specific sensor biases and recalibrates predictions accordingly. By incorporating these inferred biases, our approach generates posterior predictive distributions for engine-out NOx on unseen test data, achieving high accuracy without retraining the model. Our results demonstrate that this transferable modeling approach significantly improves the accuracy of predictions compared to conventional non-adaptive GP models, effectively addressing engine-to-engine variability and improving model generalizability.
\end{abstract}


\maketitle

\section{Introduction}
\label{sec:intro}

Accurate on-board modeling of engine-out NOx (the nitrogen oxides exiting an engine before an aftertreatment system) is critical for meeting increasingly stringent emissions regulations and improving engine performance. Engine-out NOx, primarily generated in diesel and gasoline engines during combustion, contributes significantly to air pollution, smog formation, and health issues such as respiratory diseases~\citep{boningari2016impact}. As emission regulations such as Euro VI/VII~\citep{european2022commission} and U.S. EPA Tier III/IV~\citep{epa} continue to evolve, automotive manufacturers face the challenge of reducing emissions under various operating conditions, including real driving scenarios~\citep{barbier2023predicting}. Moreover, new testing procedures such as real driving emissions tests demand that engines maintain low NOx outputs not just in laboratory cycles but during on-road transients~\citep{bajwa2024development}. 
Meeting these regulations requires precise control of in-cylinder combustion and efficient aftertreatment usage, both of which depend on accurate engine-out NOx models. 

However, achieving high fidelity NOx prediction is technically challenging. NOx formation in engines is sensitive to many factors, such as combustion temperature, oxygen availability, fuel injection timing, and exhaust gas recirculation rates, which interact in a highly nonlinear fashion. 
Traditional NOx sensors are costly and operate only when exhaust temperatures are high enough, meaning there is no knowledge of engine-out NOx during cold exhaust conditions, such as engine startup. This limitation is particularly critical because a substantial fraction of the total cumulative tailpipe NOx is emitted during this cold-start and warm-up phase, before the aftertreatment system reaches its optimal operating temperature.
This has spurred interest in predictive models that estimate engine-out NOx from readily available engine signals. 
Instead of substituting NOx sensors with a purely model based virtual sensor, model predictions can be used to improve the functionality of the physical sensors by providing a reference signal for consistency checks. 
For example, incorporating NOx prediction from the engine can improve the sensor signal integrity of physical sensors and help identify malfunctions or aging within the aftertreatment system~\citep{barbier2023predicting}. Furthermore, the estimation of real time NOx serves as a critical parameter for the vehicle's on-board diagnostics system, allowing effective monitoring of engine states~\citep{walker2016future}.

In recent years, machine learning (ML) approaches have been extensively explored~\citep{aliramezani2022modeling} for predicting engine-out NOx due to their ability to model complex, nonlinear relationships without requiring detailed physical insight into combustion phenomena. Artificial neural networks (ANNs)~\citep{fang2022artificial}, support vector machines (SVMs)~\citep{aliramezani2020support}, and deep neural networks (DNNs)~\citep{shin2020deep} have demonstrated high accuracy under both steady state and transient conditions. For instance, ~\citep{shin2020deep} designed a deep learning model to predict transient engine-out NOx using steady state datasets, achieving notable accuracy by capturing input output relationships under dynamic conditions. ~\citep{fang2022artificial} used ANNs to predict transient engine-out NOx from high speed direct injection diesel engines. Similarly, ~\citep{pillai2022modeling} developed DNN models for engine-out and tailpipe NOx prediction in heavy duty vehicles using onboard diagnostic data. ~\citep{aliramezani2020support} used SVMs to develop control-oriented NOx models, outperforming conventional regression algorithms in both transient and steady state conditions. Despite their success, these data-driven models often suffer from poor generalization when applied to unseen engine datasets and cannot respond to changes in the underlying physics of the system.

To overcome the limitations of purely data-driven approaches, physics-based and semi empirical models incorporate underlying combustion principles to predict engine-out NOx. For example, ~\citep{asprion2013fast} proposed a fast and accurate physics-based model for NOx. Their other work ~\citep{asprion2013optimisation} used simplified physical formulations to model NOx formation under real driving conditions. Their approach leveraged thermodynamic parameters such as combustion temperature and oxygen concentration to achieve computational efficiency. Similarly, ~\citep{mohammad2023physical} combined physical knowledge with dimensionality reduction to develop hybrid emission models, demonstrating that physically informed inputs improve NOx prediction accuracy. ~\citep{aithal2010modeling} used finite rate chemical kinetics to model NOx formation in diesel engines, highlighting the role of combustion processes. Another prominent study by ~\citep{lee2021real} developed real time NOx estimation models by embedding combustion parameters into a simplified physical framework. Although physics-based models are robust to variations in operating conditions, their reliance on empirical calibrations and in-cylinder measurements often limits their transferability across engines.

Hybrid approaches combine the strengths of physics-based and ML methods to improve prediction accuracy and reduce computational costs. In these models, the combustion process is initially simulated using a straightforward physical model. The output of this physical simulation is then integrated with experimental data to identify the critical inputs required for training a data-driven methodology. This approach ensures that these models remain efficient and precise due to their reliance on data-driven techniques, while also maintaining adaptability to variations in the underlying physical phenomena, as it incorporates a physical combustion model~\citep{rezaei2020hybrid}. 
In ~\citep{mohammad2021hybrid}, a hybrid framework was proposed where GT-SUITE, a 1D simulation tool, was coupled with SVMs and feedforward neural networks to predict engine-out NOx. Their approach was evaluated on 772 steady‑state operating points from a 13‑liter heavy‑duty diesel engine, achieving an $R^2$ of 0.99. However, its performance under transient conditions was not assessed.

Recent advances in physics informed neural networks~\citep{raissi2019physics, nath2023physics} have further extended hybrid modeling paradigms by embedding governing physical laws directly into neural network training through physics based constraints in the loss function. Such approaches have demonstrated improved generalization and physical consistency in engine modeling tasks by incorporating thermodynamic relationships and combustion constraints into data driven architectures. While these methods improve model consistency with underlying physics, they are typically trained in a supervised manner on engine specific datasets, and their deployment across different engines may require additional adaptation or recalibration.

Despite advances in data-driven, physics-based, and hybrid techniques, the challenge of developing engine-to-engine transferable models remains largely unaddressed. Most of these models are engine specific and fail to generalize across engines of the same type due to differences such as manufacturing imperfections and mechanical wear. Ensuring engine-to-engine transferability with minimal retuning remains an open problem in the field. Therefore, developing models that can accommodate this variability is essential for improving scalability, reducing calibration costs, and allowing robust NOx prediction across multiple engines. There have been some attempts to improve transferability. For instance, ~\citep{shin2022designing} highlighted the domain mismatch problem between steady state and transient conditions, but did not explore engine-to-engine transferability. In another study ~\citep{shin2023task} used transfer learning in which knowledge from one task or dataset is transferred to speed up learning on a new task or engine. Their results showed that reusing a pretrained model can indeed save time and still achieve accuracy comparable to training a dedicated model for the new task or engine. However, this method required retraining for each new engine, a process that is computationally expensive and impossible for large scale deployment.

Bayesian methods have shown potential to address model uncertainties and improve generalization. For example, ~\citep{cho2018structured} proposed a structured approach to uncertainty analysis of predictive models of engine-out NOx. Similarly, our previous work ~\citep{zinage2025causal} proposed a causal graph-enhanced Gaussian process (GP) regression to model engine-out NOx.
that uses a hybrid architecture using deep kernel learning~\citep{wilson2016deep, zinage2024dkl}. 
However, this work has yet to be extended to tackle sensor biases and engine-to-engine variability effectively. To bridge this gap, this paper introduces a Bayesian framework that accounts for sensor biases while leveraging a pre-trained GP model. By including engine specific parameters, our approach achieves high predictive accuracy across different engines without retraining the GP model. This innovative solution not only improves engine-to-engine transferability but also quantifies uncertainty due to variability across different engines.

We have organized our paper as follows. We first explore the fundamental physical principles that govern NOx formation. This is followed by the problem statement before discussing the methodology. 
Next, we present the experimental setup followed by results of our study. Finally, we conclude with a concise summary of our findings. This research was carried out in collaboration with Cummins Inc., with data coming from Cummins medium-duty diesel engines. In accordance with Cummins policies, all plots presented in this research have been normalized.

\section{Engine-out NOx formation}
\label{sec:engine_out_nox}

\begin{figure}[H]
    \centering
    \includegraphics[width=0.9\linewidth]{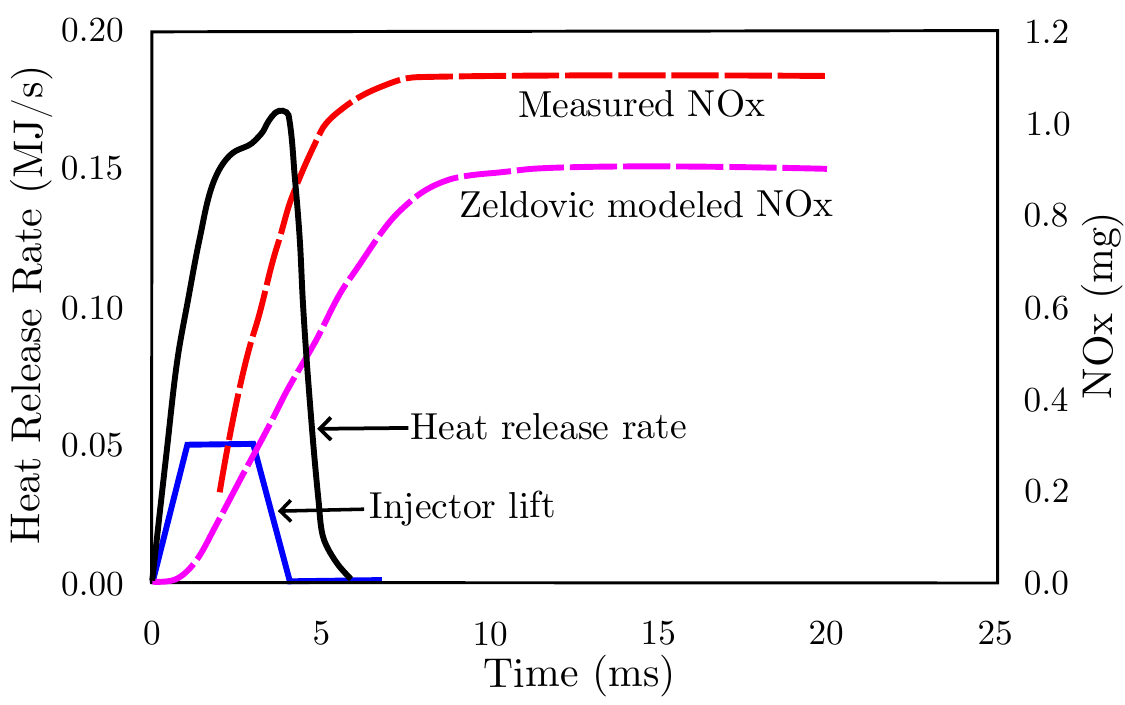}
    \caption{Formation of NO\textsubscript{x} in direct injected diesel compression ignition engines (Source: Figure adapted from \cite{khair2015nox})}
    \label{fig:nox_formation}
\end{figure}

The formation of nitrogen oxides (NO\textsubscript{x}) (Fig. \ref{fig:nox_formation}), primarily comprising nitric oxide (NO) and nitrogen dioxide (NO\textsubscript{2}), in diesel compression ignition engines is a complex phenomenon governed by multiple chemical pathways. The well recognized pathways include thermal NO\textsubscript{x}, prompt NO\textsubscript{x}, and fuel bound NO\textsubscript{x}. Among these, the thermal NO\textsubscript{x} pathway, described by the extended Zeldovich mechanism, is the most significant contributor to engine-out NO\textsubscript{x} emissions in diesel engines. 

Thermal NO\textsubscript{x} arises from the high temperature reaction of atmospheric nitrogen (N\textsubscript{2}) with oxygen. It is described by the extended Zeldovich mechanism~\citep{lavoie1970experimental}, which involves the following key reactions:
\begin{equation*}
\begin{split}
\mathrm{N}_2 + \mathrm{O} \rightarrow \mathrm{NO} + \mathrm{N}, \\
\mathrm{N} + \mathrm{O}_2 \rightarrow \mathrm{NO} + \mathrm{O}, \\
\mathrm{N} + \mathrm{OH} \rightarrow \mathrm{NO} + \mathrm{H}.
\end{split}
\end{equation*}
These reactions are highly sensitive to the in-cylinder temperature, typically becoming significant above 2000 K~\citep{bowman1975kinetics}. Additionally, the concentration of oxygen (O\textsubscript{2}) and the residence time at peak combustion temperatures in lean air fuel mixtures significantly influence the rate of thermal NO formation. 
Prompt NO\textsubscript{x}, also known as the Fenimore mechanism, refers to NO formed rapidly in the fuel rich flame front before thermal equilibrium is reached. This pathway is relatively less dependent on temperature and plays a minor role in diesel engines due to the predominantly lean combustion conditions. The classic prompt NO initiation reaction is
\begin{equation*}
  \mathrm{CH} + \mathrm{N}_2 \rightarrow \mathrm{HCN} + \mathrm{N}. \\  
\end{equation*}
This pathway is favored in fuel rich combustion regions where hydrocarbon radicals are abundant. It occurs on the timescale of the flame propagation, much faster than thermal NO formation.

Fuel bound NO\textsubscript{x} is generated by the oxidation of nitrogen present in the fuel itself. If the fuel contains organically bound nitrogen, this nitrogen can be released during combustion and subsequently oxidized into NO or NO\textsubscript{2}. This mechanism is well known to be dominant in combustion of high nitrogen fuels such as coal or heavy fuel oil. In such cases, essentially all the fuel bound nitrogen can end up as NO\textsubscript{x} in the exhaust. However, conventional diesel fuels have very low nitrogen content, so fuel NOx is negligible in most diesel engines. The same holds for biodiesel, which generally contains minimal fuel nitrogen. 

Thus, while the fuel NOx mechanism exists in theory, it is not a significant NOx source for diesel engines running on ultra low sulfur diesel or biodiesel. The conversion of NO to NO\textsubscript{2} occurs predominantly downstream of the combustion chamber, where conditions favor incomplete oxidation of NO. The key reaction~\citep{merryman1975nitrogen} involved in this process is: 
\begin{equation*}
\mathrm{NO} + \mathrm{HO}_2 \rightarrow \mathrm{NO}_2 + \mathrm{OH}. 
\end{equation*} 
Although NO\textsubscript{2} constitutes a smaller fraction of the total NO\textsubscript{x} emissions, its environmental and health impacts necessitate its consideration in emission studies. The levels of engine-out NO\textsubscript{x} emissions are primarily governed by the combustion temperature, the oxygen concentration in the combustion chamber, and the duration of high temperature exposure. These parameters are, in turn, influenced by engine operating conditions, including intake air mass flow rate, fuel injection parameters and engine speed/load. A widely adopted technique for reducing engine-out NO\textsubscript{x} emissions is exhaust gas recirculation (EGR). EGR involves recirculating a portion of the exhaust gases, predominantly composed of nitrogen (N\textsubscript{2}), carbon dioxide (CO\textsubscript{2}), and water vapor (H\textsubscript{2}O), back into the intake air stream. The introduction of these inert gases serves to lower oxygen concentration and reduce combustion temperature. EGR implementation requires careful optimization to balance NO\textsubscript{x} reduction with potential adverse effects such as increased particulate matter emissions and reduced engine efficiency. Recent advancements in EGR technology, including cooled EGR and variable rate EGR systems, have improved the effectiveness and adaptability of this approach across diverse operating conditions.

\section{Problem Statement}
\label{sec:problem_statement}

Accurately predicting engine-out NOx across a population of nominally identical engines remains unresolved due to sensor biases, engine-to-engine variability, and shifts in operating conditions that degrade the performance of conventional data-driven and physics-based models. Existing ML approaches achieve high accuracy on the engines used for training but fail to generalize to new engines without extensive recalibration or full retraining, which is impractical for large scale deployment. 

The core problem addressed in this work is to develop a Bayesian calibration framework for engine-out NOx models that simultaneously (i) estimates sensor biases intrinsic to individual engines, (ii) adapts a pre-trained model to new engine using only a small amount of engine-specific data and (iii) produces reliable posterior predictive distributions (PPDs) that quantify uncertainty due to engine-to-engine variability. 

\section{Methodology}
\label{sec:methodology}

\begin{figure*}[htbp]
    \centering
    \includegraphics[width = 16.8cm]{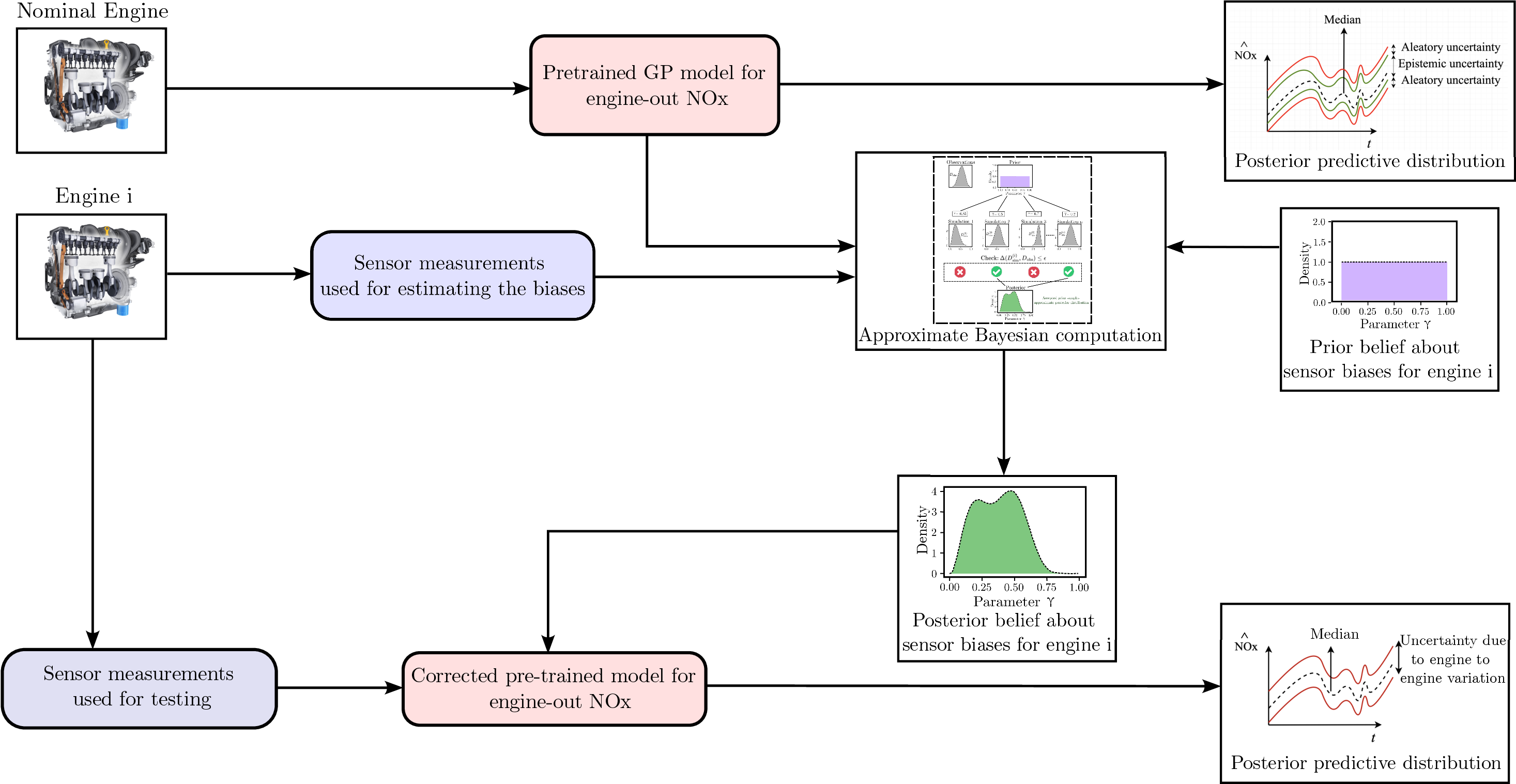}
    \caption{Bayesian calibration framework for engine-to-engine transferability.}
    \label{fig:bayesian_calibration_framework}
\end{figure*}

We propose a Bayesian calibration framework, illustrated in Fig.~\ref{fig:bayesian_calibration_framework}, designed to estimate engine-specific sensor biases from limited experimental data obtained from individual engines, alongside a pre-trained GP model derived from nominal engine data. The inferred biases are then integrated into this predictive model to allow robust, probabilistic predictions of engine-out NOx without retraining of the underlying GP.

The pre-trained GP model maps a vector of engine and combustion inputs to engine-out NOx, using a temporal input window of 5 seconds ($W_s = 5\,\mathrm{s}$) to effectively capture dynamic memory effects. We denote this model as GP (RBF) [$W_s = 5\,\mathrm{s}$], as this particular configuration previously exhibited superior predictive performance overall compared to alternative GP variants reported in \cite{zinage2025causal}.

The input vector $x_{i,t} \in \mathbb{R}^{d}$ comprises the following variables:
\begin{enumerate}
\item Turbine inlet temperature
\item Engine speed
\item EGR valve actuation
\item VGT valve actuation
\item Mass flow rate of EGR
\item Mass flow rate of air
\item Fuel rail pressure
\item Engine brake torque
\item Main injection timing
\item Main injection quantity
\item Pilot 2 injection timing
\item Pilot 2 injection quantity
\item Post 1 injection timing
\item Post 1 injection quantity
\item In-cylinder O$_2$ concentration
\item EGR system outlet temperature
\item Intercooler outlet temperature
\end{enumerate}

\begin{figure}[htbp]
    \centering
    \begin{subfigure}[b]{0.99\linewidth}
        \centering
        \includegraphics[width=\linewidth]{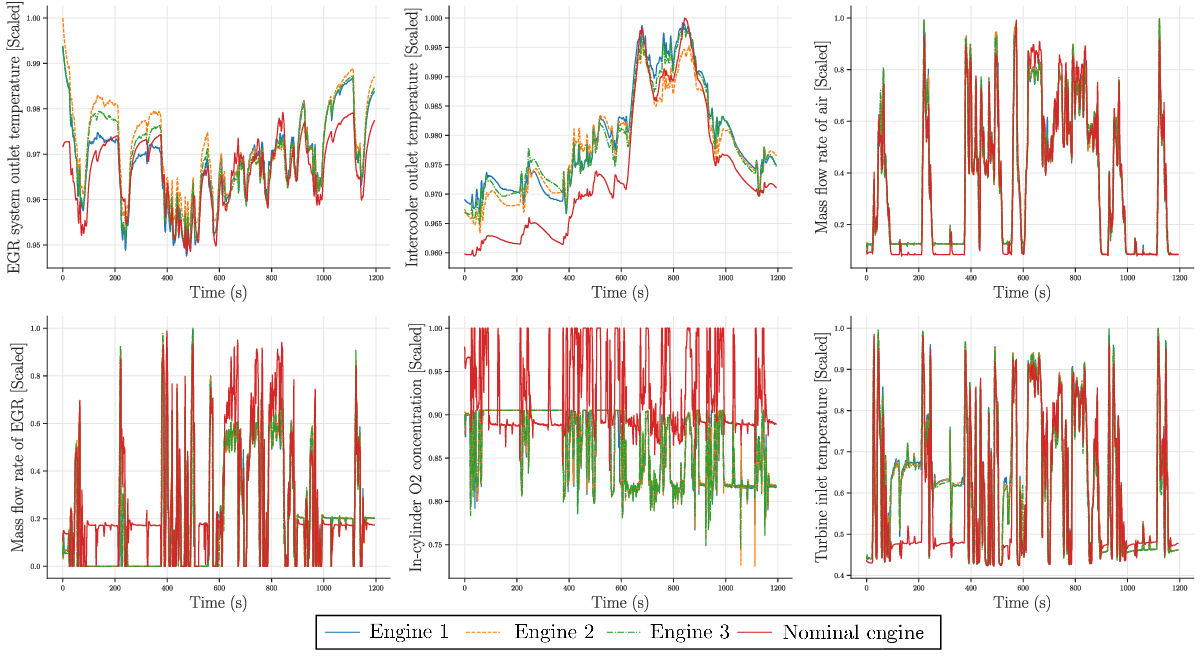}
        \caption{FTP cycle}
        \label{fig:inputs_ftp_cycle}
    \end{subfigure}
    
    \vspace{0.5cm} 

    \begin{subfigure}[b]{0.99\linewidth}
        \centering
        \includegraphics[width=\linewidth]{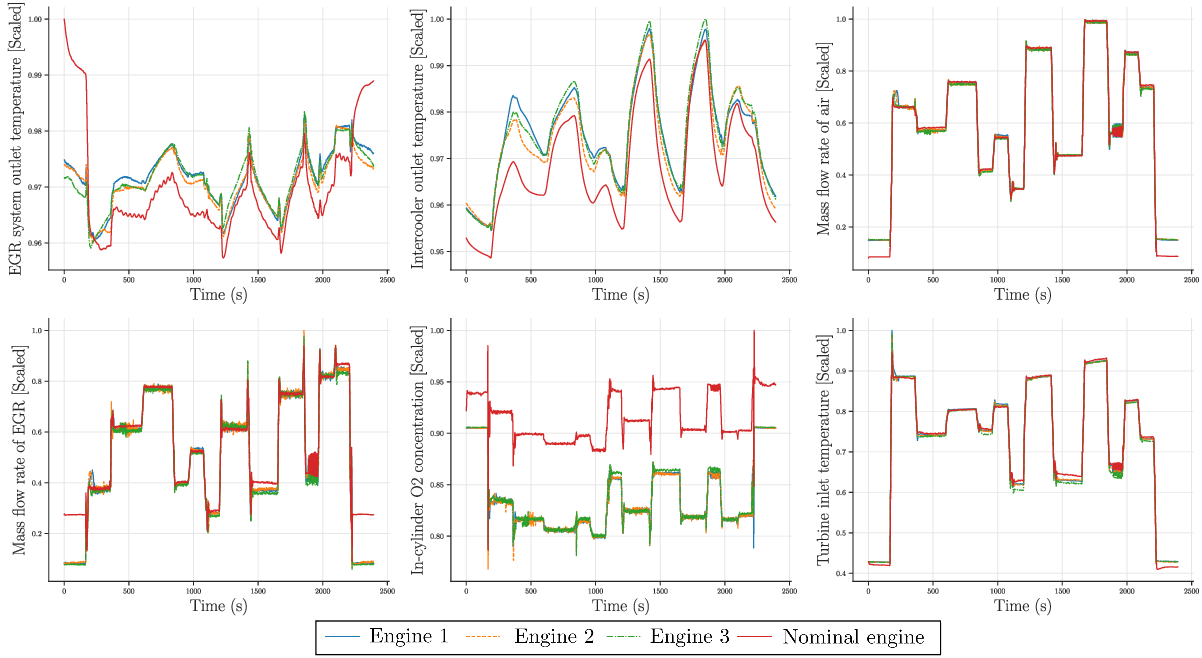}
        \caption{SET cycle}
        \label{fig:inputs_rmcset_cycle}
    \end{subfigure}
    \caption{Sensor measurements for non-control input variables across different engines}
    \label{fig:inputs_cycle}
\end{figure}

Fig.~\ref{fig:inputs_cycle} illustrates the variations in sensor measurements across non-control input variables due to calibration discrepancies when transitioning between engines. It is thus reasonable to assume sensor biases remain constant across time steps, as consistent offsets can be observed across engines.

Formally, let $d_{\text{nc}}$ denote the number of non-control (measured) input variables. For engine $i$, we define the sensor bias vector as $b_i \in \mathbb{R}^{d_{\text{nc}}}$, assumed time invariant. To map these biases onto the full input vector, we introduce a fixed selection matrix $S \in \{0,1\}^{d \times d_{\text{nc}}}$, structured such that it selects only non-control inputs, leaving control inputs unchanged. For instance, if the input vector $x$ is partitioned as $x = [u_{\text{control}}, x_{\text{measured}}]^T$, then $S$ takes the block form $S = [0, I_{d_{\text{nc}}}]^T$. Here, $I_{d_{\text{nc}}}$ is the identity matrix corresponding to measured inputs (such as EGR system outlet temperature, intercooler outlet temperature, turbine inlet temperature) which are subject to sensor bias, and the zero matrix ensures that control inputs (such as VGT valve actuation, EGR valve actuation) remain uncorrected.
The bias corrected input vector used for prediction is thus defined as:
\begin{equation*}
\tilde{x}_{i,t} = x_{i,t} - S b_i \in \mathbb{R}^{d}.
\end{equation*}

We denote $y_{i,t}$ as the observed engine-out NOx measurement for engine $i$ at time $t$, and let $y_{i,1:T} = \{y_{i,t}\}_{t=1}^T$ and $x_{i,1:T} = \{x_{i,t}\}_{t=1}^T$ represent sequences of engine-out NOx and inputs respectively over a time horizon $T$. The GP model $f(\cdot)$ operates in a normalized space defined by a transformation $Q(\cdot)$ where predictions are subsequently mapped back to physical NOx units via the inverse transformation $Q^{-1}(\cdot)$. To simplify notation, we define the GP based median predictor in the physical NOx scale as
\[
    g(x) = \operatorname{MEDIAN}\!\left(
        Q^{-1}\!\left(
        f\!\left(Q(x)\right)
        \right)
    \right),
\]
where the median is taken with respect to the GP predictive distribution and is used instead of the mean because the inverse transformation $Q^{-1}(\cdot)$ maps the GP outputs to a highly skewed, non-negative distribution in the physical domain.
\begin{figure}[htbp]
    \centering
    \begin{subfigure}[b]{0.9\linewidth}
        \centering
        \includegraphics[width=\linewidth]{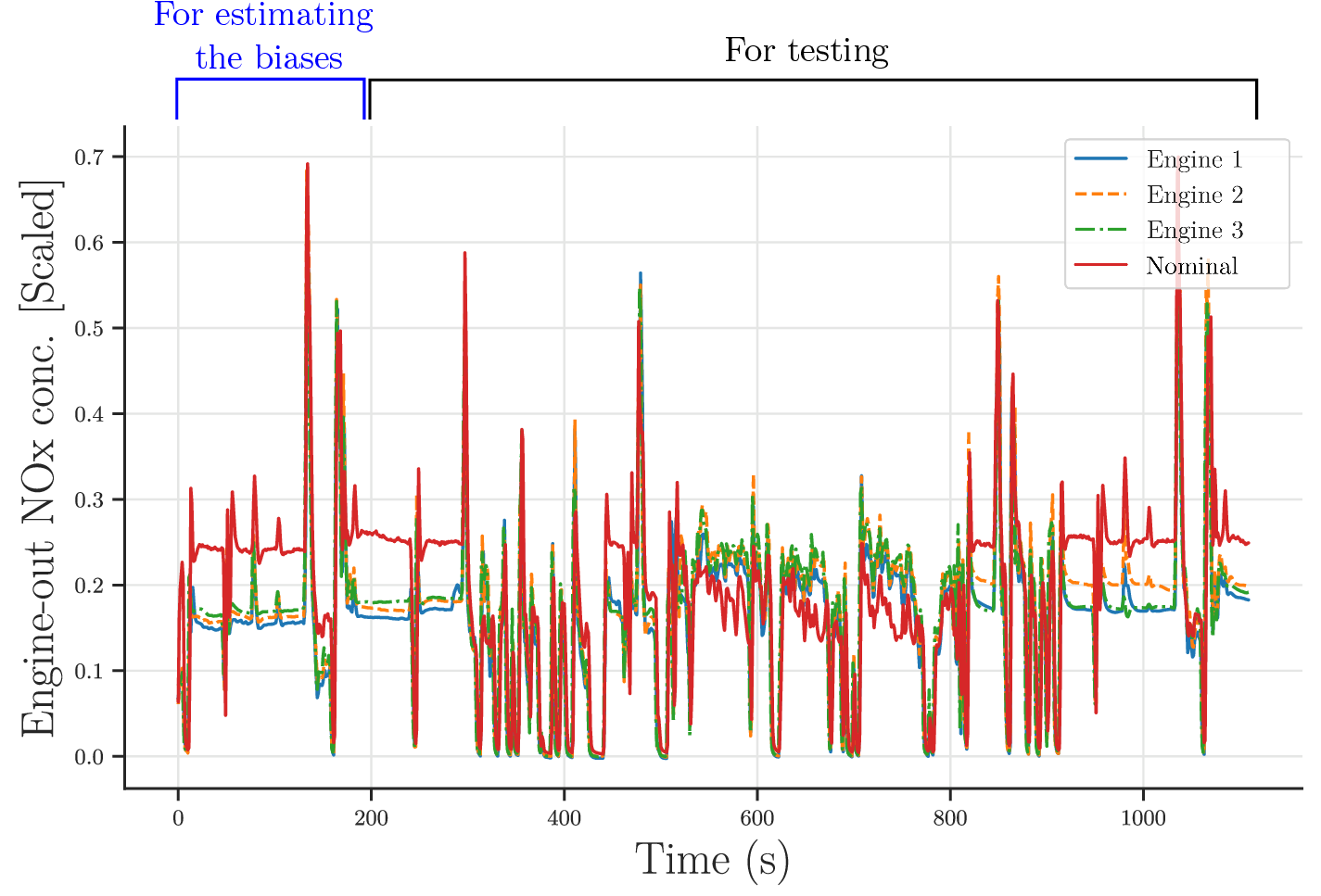}
        \caption{FTP cycle}
        \label{fig:nox_ftp_cycle_different_engines_1}
    \end{subfigure}%
    \hfill
    \begin{subfigure}[b]{0.9\linewidth}
        \centering
        \includegraphics[width=\linewidth]{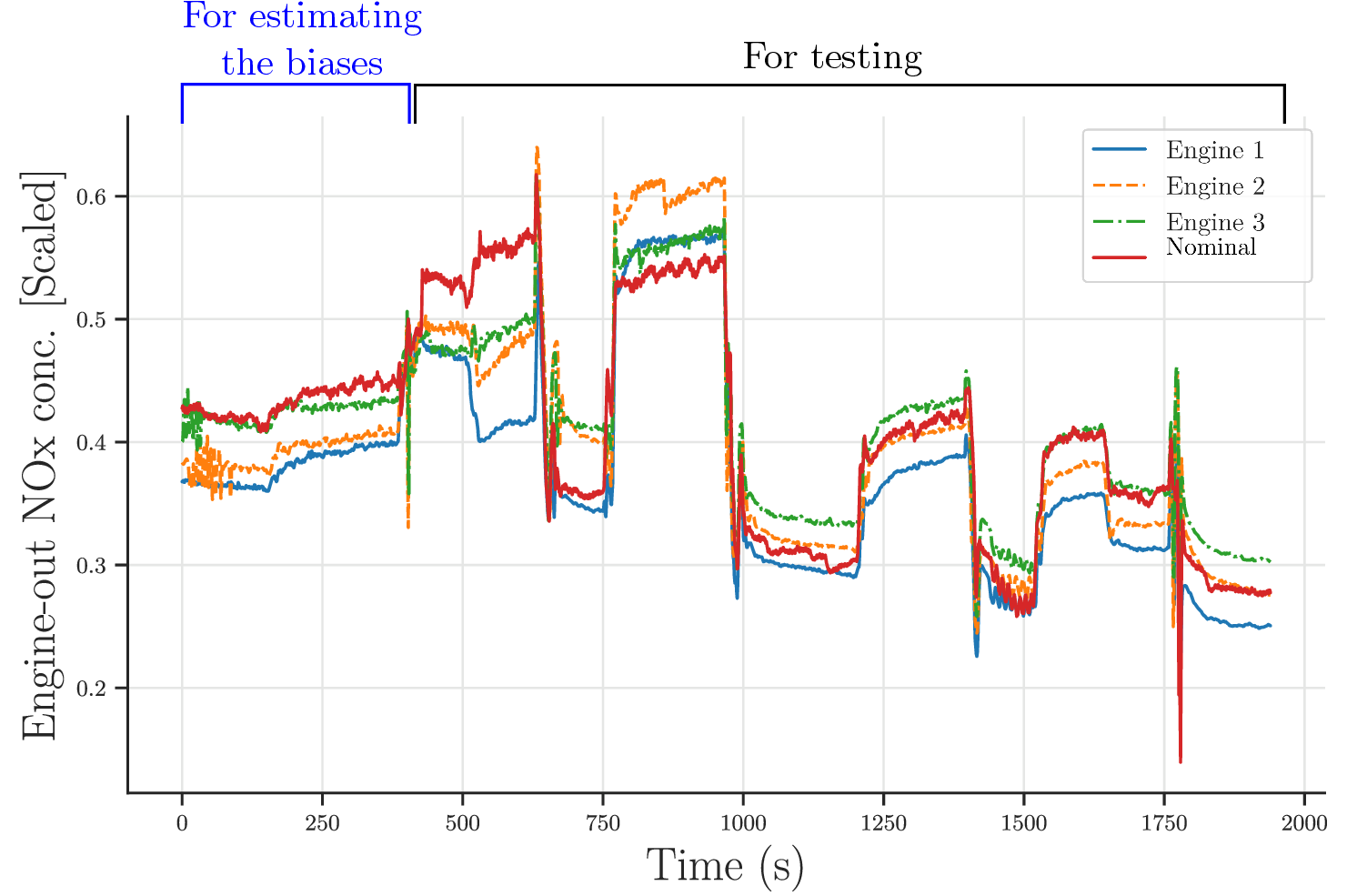}
        \caption{SET cycle}
        \label{fig:nox_set_cycle_different_engines_1}
    \end{subfigure}
    \caption{Variation in NOx measurements across different engines}
    \label{fig:nox_different_engines}
\end{figure}

In such heavy tailed distributions, the mean is often pulled toward extreme values, whereas the median remains a robust estimator of the central tendency and is invariant under the monotonic transformation $Q^{-1}$.
The measurement model for engine-out NOx is then expressed as:
\begin{equation}
y_{i,t} = g(\tilde{x}_{i,t}) + \alpha_i + \varepsilon_{i,t}, \quad \varepsilon_{i,t} \sim \mathcal{N}(0, \sigma_y^2),
\label{eq:measurement_model}
\end{equation}
where $\alpha_i \in \mathbb{R}$ denotes an additive bias specific to engine $i$ which remains constant over time (as supported by empirical observations in Fig.~\ref{fig:nox_different_engines}), and $\sigma_y^2$ represents the observation noise variance, assumed homoscedastic and temporally independent. Consequently the complete set of biases for engine $i$ is represented by the parameter tuple $(\alpha_i, b_i)$.

Assuming conditional independence of observations over time given $(\alpha_i, b_i)$ and the GP based median predictor $g(\cdot)$, the likelihood function for engine $i$ is:
\begin{equation}
\scalebox{0.8}{$
\begin{aligned}
p(y_{i,1:T} \mid \alpha_i, b_i, x_{i,1:T})
&= \prod_{t=1}^T
\mathcal{N}\!\left(
y_{i,t} \mid g(x_{i,t} - S b_i) + \alpha_i,
\sigma_y^2
\right).
\end{aligned}
$}
\label{eq:likelihood}
\end{equation} 
The priors for the biases are chosen to be uniform distributions, with bounds selected using domain knowledge provided by Cummins regarding the expected ranges of sensor biases. These bounds encode practical engineering constraints on plausible measurement discrepancies between the nominal and sample engines. The joint prior thus factorizes as:
\begin{equation*}
p(\alpha_i, b_i) = p(\alpha_i) \prod_{k=1}^{d_{\text{nc}}} p\left(b_i^{(k)}\right).
\end{equation*}
Integrating the likelihood in Eq. \ref{eq:likelihood} with this prior formulation yields the posterior distribution of bias parameters for engine $i$ as:
\begin{equation*}
\scalebox{0.95}{$
\begin{aligned}
p(\alpha_i, b_i \mid y_{i,1:T}, x_{i,1:T})
&\propto
p(y_{i,1:T} \mid \alpha_i, b_i, x_{i,1:T}) p(\alpha_i)p(b_i).
\end{aligned}
$}
\end{equation*}
We use an approximate Bayesian Computation (ABC) approach, detailed in Algorithm~\ref{alg:abc_rejection_nox}, to approximate the posterior distribution $p(\alpha_i, b_i \mid y_{i,1:T}, x_{i,1:T})$. ABC is a likelihood free inference technique that generates parameter samples from the prior and selects those that produce simulated data sufficiently close to observed data according to a predefined distance metric. Additional methodological details and theoretical foundations of ABC are provided in the appendix.

\begin{algorithm}[htbp]
\caption{Bayesian calibration of engine-out NOx models for engine-to-engine transferability}
\begin{algorithmic}
\Require Pre-trained GP based median predictor $g(\cdot)$, engine dataset $D_{\text{obs}}^{(i)}=\{x_{i,1:T}, y_{i,1:T}\}$, selection matrix $S$, priors $p(\alpha_i)$ and $p(b_i)$, observation noise variance $\sigma_y^2$, pilot draws $N_{\text{pilot}}$, main draws $N_{\text{main}}$, desired accepted samples $N_{\text{desired}}$, Quantile $\zeta\in(0,1)$, distance metric $\Delta$, new input sequence $x_{\ast}$ for prediction
\Ensure Posterior sample set $\mathcal{S}^{(i)}$
    \Statex
    \State \textbf{Pilot sampling phase to select the tolerance}
    \State $\mathcal{D}_{\text{pilot}}\gets \varnothing$
    \For{$s \gets 1$ \textbf{to} $N_{\text{pilot}}$}
        \State Draw $\alpha_i^{(s)} \sim p(\alpha_i)$ and $b_i^{(s)} \sim p(b_i)$
        \State $\tilde{x}^{(s)} \gets \{x_{i,t} - S b_i^{(s)}\}_{t=1}^T$ 
        \State $y_{\text{det}}^{(s)} \gets \{g(\tilde{x}_t^{(s)}) + \alpha_i^{(s)}\}_{t=1}^T$ 
        \State Draw $Z \sim \mathcal{N}_{T}(0,I)$
        \State $y_{\text{sim}}^{(s)} \gets y_{\text{det}}^{(s)} + \sigma_y \cdot Z$ 
        \State $d_s \gets \Delta(y_{\text{sim}}^{(s)}, y_{i,1:T})$
        \State append $d_s$ to $\mathcal{D}_{\text{pilot}}$
    \EndFor
    \State $\epsilon_{\text{ABC}}\gets$\texttt{Quantile}({$\mathcal{D}_{\text{pilot}},\zeta$})
    \Statex
    \State \textbf{Main sampling phase with the fixed tolerance}
    \State $\mathcal{S}^{(i)} \gets \varnothing$
    \For{$s \gets 1$ \textbf{to} $N_{\text{main}}$}
        \State Draw $\theta^{(s)}=[\alpha_i^{(s)}; b_i^{(s)}]$ from priors $p(\alpha_i), p(b_i)$
        \State $\tilde{x}^{(s)} \gets \{x_{i,t} - S b_i^{(s)}\}_{t=1}^T$
        \State $y_{\text{det}}^{(s)} \gets \{g(\tilde{x}_t^{(s)}) + \alpha_i^{(s)}\}_{t=1}^T$
        \State Draw $Z \sim \mathcal{N}_{T}(0,I)$
        \State $y_{\text{sim}}^{(s)} \gets y_{\text{det}}^{(s)} + \sigma_y \cdot Z$
        \State $d_s \gets \Delta(y_{\text{sim}}^{(s)}, y_{i,1:T})$
        \If{$d_s \le \epsilon_{\text{ABC}}$}
            \State $\mathcal{S}^{(i)} \gets \mathcal{S}^{(i)} \cup \{\theta^{(s)}\}$
            \If{$|\mathcal{S}^{(i)}| = N_{\text{desired}}$}
                \State \textbf{break}
            \EndIf
        \EndIf
    \EndFor
    \Statex
    \State \textbf{Posterior predictive generation for new input} $x_{\ast}$
    \State $\mathcal{Y}_{\text{pred}} \gets \varnothing$
    \For{$\theta^{(s)} \in \mathcal{S}^{(i)}$}
        \State Parse $\theta^{(s)}$ into $\alpha_i^{(s)}$ and $b_i^{(s)}$
        \State $\tilde{x}_{\ast}^{(s)} \gets \{x_{\ast,t} - S b_i^{(s)}\}_{t=1}^{T_{\ast}}$ 
        \State Draw $Z_{\ast} \sim \mathcal{N}_{T_{\ast}}(0,I)$
        \State $y_{\ast}^{(s)} \gets \{g(\tilde{x}_{\ast,t}^{(s)}) + \alpha_i^{(s)}\}_{t=1}^{T_{\ast}} + \sigma_y \cdot Z_{\ast}$
        \State $\mathcal{Y}_{\text{pred}} \gets \mathcal{Y}_{\text{pred}} \cup \{y_{\ast}^{(s)}\}$
    \EndFor
\end{algorithmic}
\label{alg:abc_rejection_nox}
\end{algorithm}

Unlike conventional ABC methods which uses fixed acceptance thresholds, our approach dynamically determines the acceptance threshold, improving efficiency and adaptively concentrating computational effort in regions of higher posterior probability. This adaptability addresses challenges inherent in high dimensional, complex problems where fixed thresholds often lead to excessive rejection rates or insufficient accuracy in posterior estimation.

The proposed ABC algorithm proceeds in two stages: a pilot phase for adaptive threshold selection and a main sampling phase to collect posterior samples. Initially, we draw $N_{\text{pilot}}$ prior samples:
\[
\{(\alpha_i^{(s)}, b_i^{(s)})\}_{s=1}^{N_{\text{pilot}}} \sim p(\alpha_i) \, p(b_i).
\]
For each sample $(\alpha_i^{(s)}, b_i^{(s)})$, we generate simulated NOx trajectories:
\begin{equation}
\begin{aligned}
y_{i,t}^{(s)}
&= g(x_{i,t} - S b_i^{(s)}) + \alpha_i^{(s)} + \varepsilon_{i,t}^{(s)}, \\
\varepsilon_{i,t}^{(s)}
&\sim \mathcal{N}(0, \sigma_y^2),\quad t = 1,\dots,T.
\end{aligned}
\label{eq:simulated_traj}
\end{equation}
The discrepancy between observed and simulated trajectories is quantified using the Kolmogorov–Smirnov (KS) statistic:
\begin{equation}
d^{(s)} = D_{n,m}^{(s)} = \sup_{t} |\hat{F}_{\text{obs}}(t) - \hat{F}_{\text{sim}}^{(s)}(t)|,
\label{eq:ks_distance}
\end{equation}
where $\hat{F}_{\text{obs}}(\cdot)$ and $\hat{F}_{\text{sim}}^{(s)}(\cdot)$ denote empirical cumulative distribution functions (ECDFs) of observed and simulated NOx values, respectively. The appendix justifies the choice for selecting KS statistic as the distance metric.

Using distances $\{d^{(s)}\}_{s=1}^{N_{\text{pilot}}}$, we determine the adaptive acceptance threshold $\epsilon_{\text{ABC}}$ as the $\zeta$-th percentile for a chosen quantile $\zeta \in (0,1)$. This procedure ensures the acceptance criterion targets parameter regions generating simulated distributions close to observed data.

In the main phase, additional prior samples $(\alpha_i^{(s)}, b_i^{(s)})$ are drawn and corresponding simulated NOx trajectories generated according to Eq.~\ref{eq:simulated_traj}. Each simulated trajectory's discrepancy $d^{(s)}$ is computed using Eq.~\ref{eq:ks_distance}, with samples accepted if and only if $d^{(s)} < \epsilon_{\text{ABC}}$. The accepted samples provide an empirical approximation to the posterior:
\begin{equation*}
\scalebox{0.9}{$
\begin{aligned}
p_{\text{ABC}}(\alpha_i, b_i \mid y_{i,1:T}, x_{i,1:T})
&\propto p(\alpha_i)\, p(b_i) \\
&\quad\times \int \mathbb{I}\!\left(d(y_{i,1:T}, y_{i,1:T}^{\text{sim}}) < \epsilon_{\text{ABC}}\right) \\
&\quad\times p\!\left(y_{i,1:T}^{\text{sim}} \mid \alpha_i, b_i, x_{i,1:T}\right)\, dy_{i,1:T}^{\text{sim}},
\end{aligned}
$}
\label{eq:abc_posterior}
\end{equation*}
where $\mathbb{I}(\cdot)$ denotes the indicator function, and $p(y_{i,1:T}^{\text{sim}} \mid \alpha_i, b_i, x_{i,1:T})$ follows directly from the generative model defined in Eq.~\ref{eq:measurement_model}.

Now let $\{(\alpha_i^{(s)}, b_i^{(s)})\}_{s=1}^{N_{\text{acc}}}$ denote the accepted ABC samples for engine $i$. Given a new input $x_{\ast}$, we construct bias corrected inputs:
\[
\tilde{x}_{\ast}^{(s)} = x_{\ast} - S b_i^{(s)},
\]
and generate posterior predictive samples:
\begin{equation*}
\begin{aligned}
y_{\ast}^{(s)} &= g(\tilde{x}_{\ast}^{(s)}) + \alpha_i^{(s)} + \varepsilon_{\ast}^{(s)}, \\
\varepsilon_{\ast}^{(s)} &\sim \mathcal{N}(0,\sigma_y^2),\quad s = 1,\dots, N_{\text{acc}}.
\end{aligned}
\label{eq:posterior_predictive_sample}
\end{equation*}
The empirical distribution of $\{y_{\ast}^{(s)}\}_{s=1}^{N_{\text{acc}}}$ approximates the PPD $p(y_{\ast} \mid x_{\ast}, y_{i,1:T}, x_{i,1:T})$, explicitly incorporating uncertainties from sensor biases thereby allowing robust and transferable NOx predictions.

\section{Experimental Setup}
\label{sec:exp_setup}

\begin{figure*}[htbp]
    \centering
    \begin{subfigure}[b]{0.32\linewidth}
        \centering
        \includegraphics[width=\linewidth]{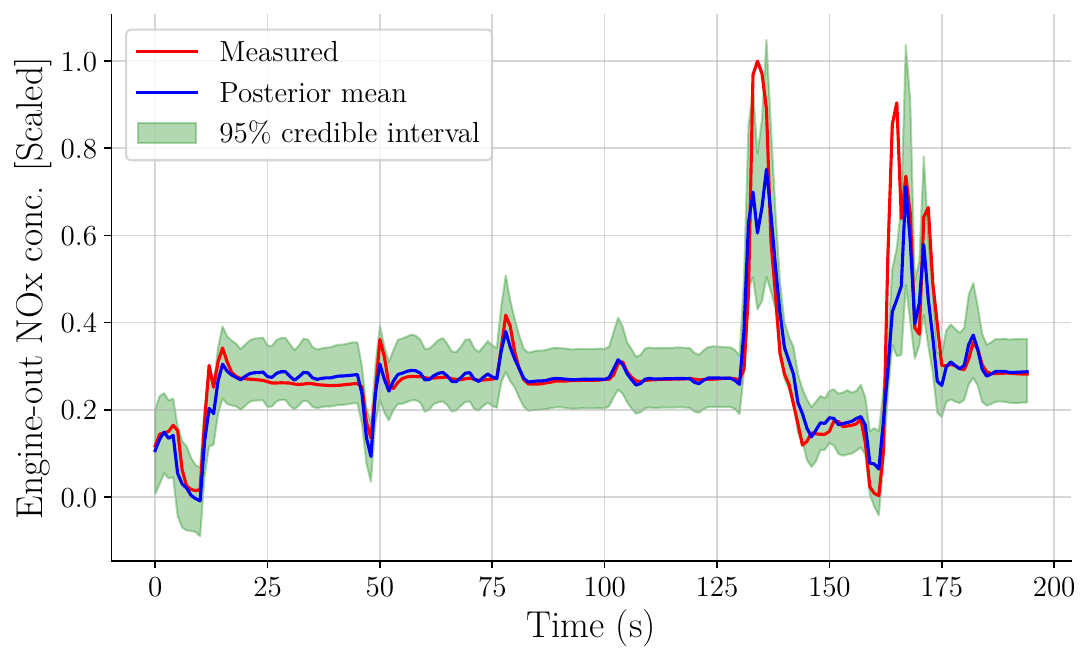}
        \caption{Engine 1}
        \label{fig:posterior_predictive_train_engine1_ftp}
    \end{subfigure}%
    \hfill
    \begin{subfigure}[b]{0.32\linewidth}
        \centering
        \includegraphics[width=\linewidth]{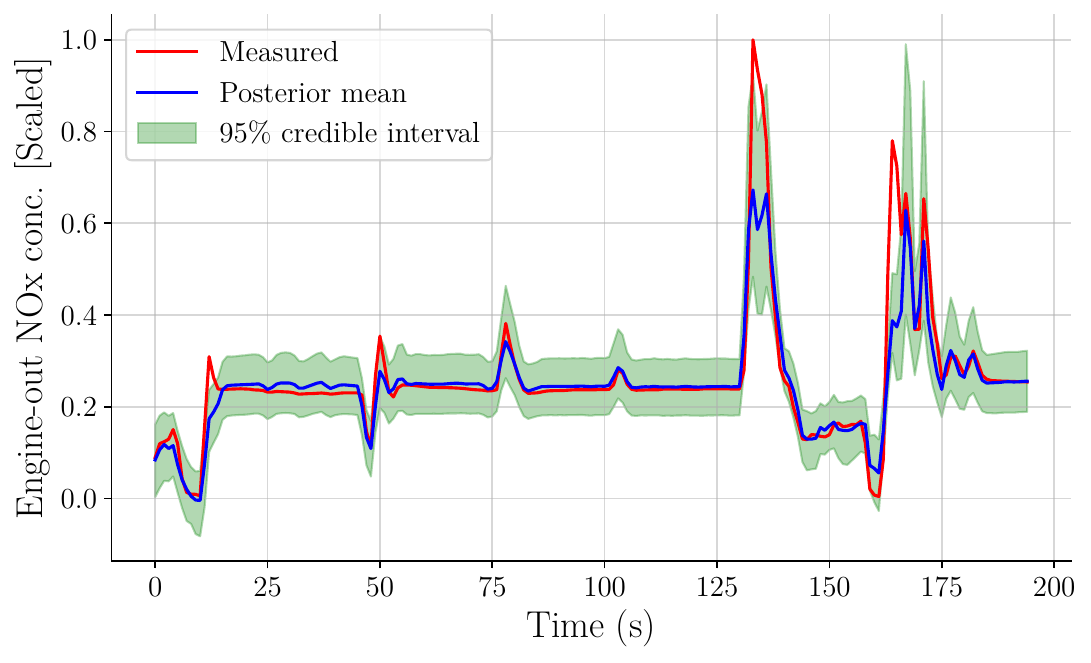}
        \caption{Engine 2}
        \label{fig:posterior_predictive_train_engine2_ftp}
    \end{subfigure}%
    \hfill
    \begin{subfigure}[b]{0.32\linewidth}
        \centering
        \includegraphics[width=\linewidth]{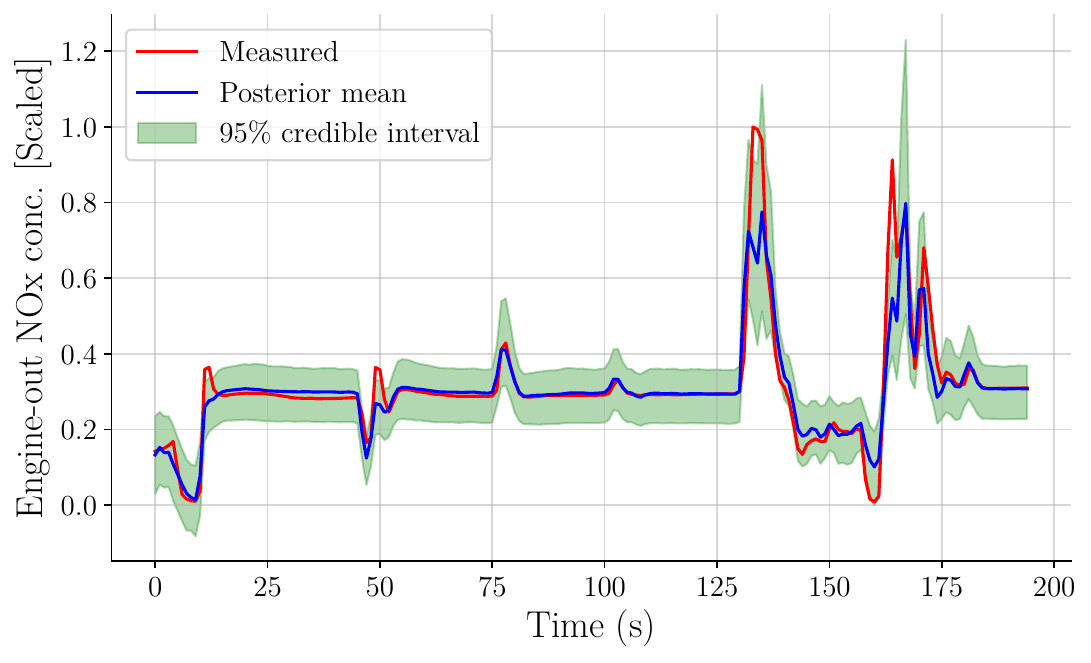} 
        \caption{Engine 3} 
        \label{fig:posterior_predictive_train_engine3_ftp}
    \end{subfigure}
    \caption{PPD of engine-out NOx on data used from FTP cycle to infer/estimate the sensor biases.}
    \label{fig:posterior_predictive_train_ftp}
\end{figure*}

\begin{figure*}[htbp]
    \centering
    \begin{subfigure}[b]{0.32\linewidth}
        \centering
        \includegraphics[width=\linewidth]{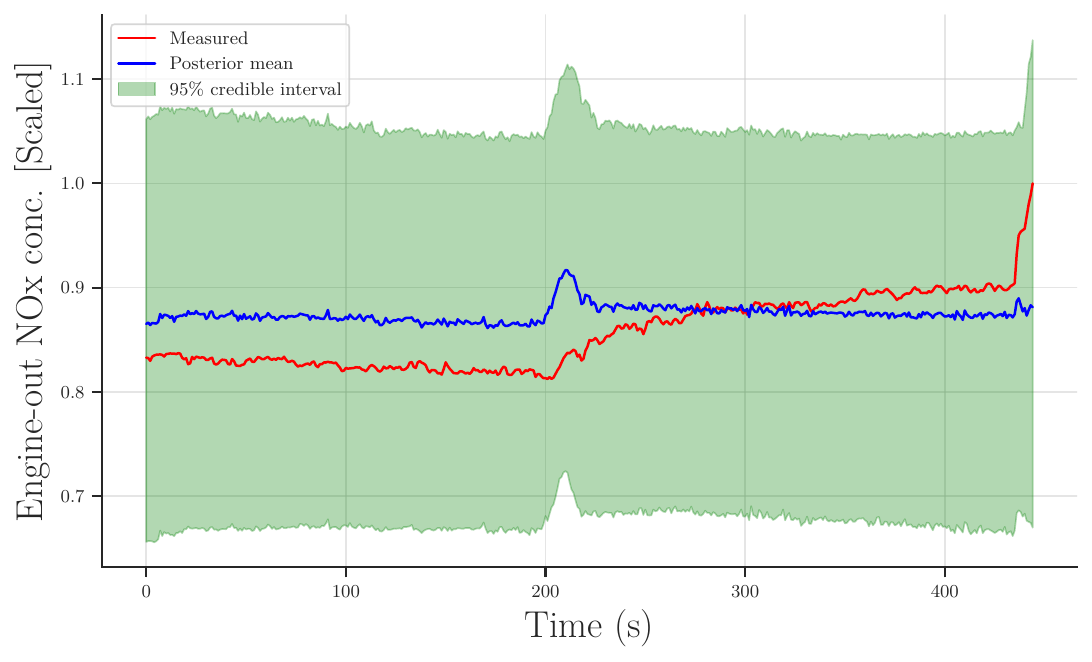}
        \caption{Engine 1}
        \label{fig:posterior_predictive_train_engine1_rmcset}
    \end{subfigure}%
    \hfill
    \begin{subfigure}[b]{0.32\linewidth}
        \centering
        \includegraphics[width=\linewidth]{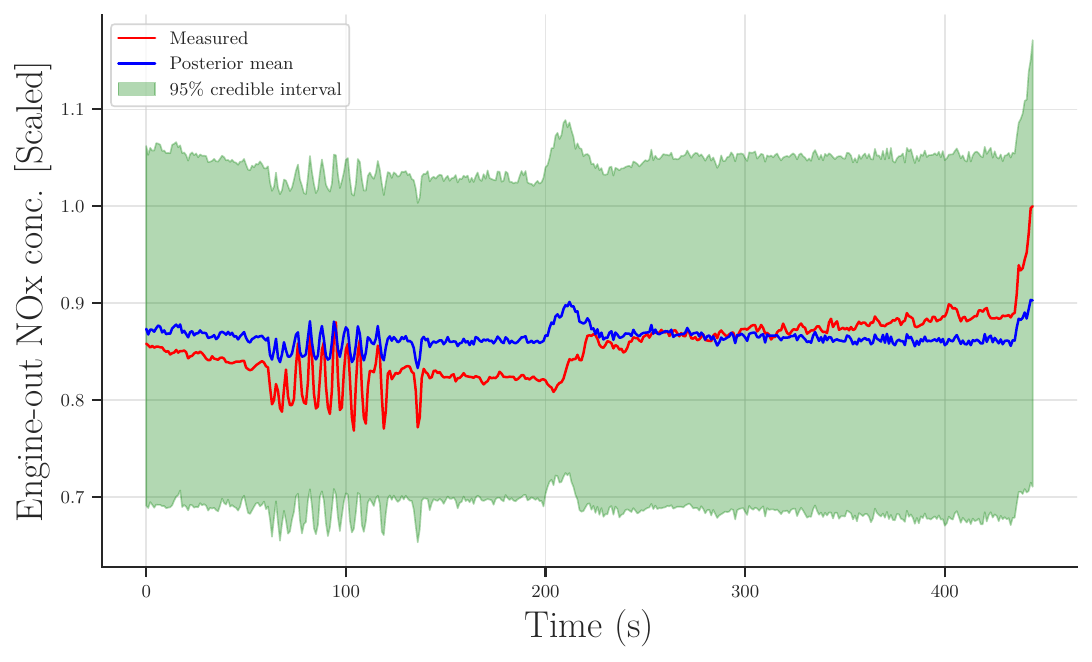}
        \caption{Engine 2}
        \label{fig:posterior_predictive_train_engine2_rmcset}
    \end{subfigure}%
    \hfill
    \begin{subfigure}[b]{0.32\linewidth}
        \centering
        \includegraphics[width=\linewidth]{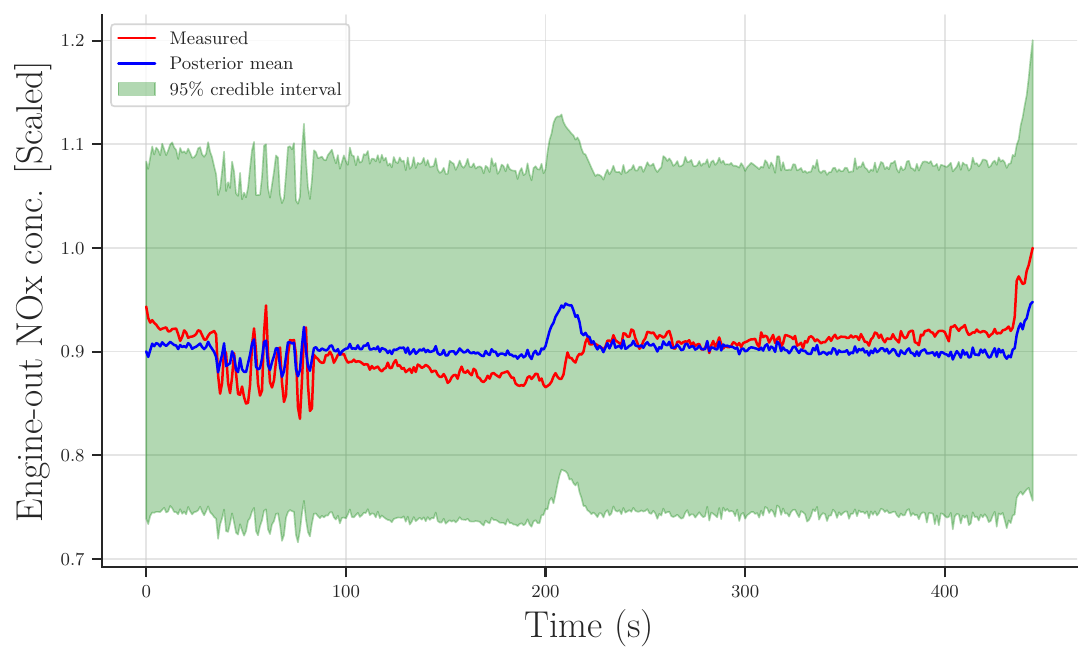} 
        \caption{Engine 3} 
        \label{fig:posterior_predictive_train_engine3_rmcset}
    \end{subfigure}
    \caption{PPD of engine-out NOx on data used from SET cycle to infer/estimate the sensor biases.}
    \label{fig:posterior_predictive_train_rmcset}
\end{figure*}

\subsection{Data Generation}

The experimental datasets used in this study were provided by Cummins Inc. and consist of measurements from four multi-pulse fueling diesel compression ignition engines. All engines belong to the Cummins B6.7L inline 6-cylinder diesel platform equipped with a high-pressure common rail fuel injection system, high-pressure EGR, and a VGT. The engines operate on ultra-low sulfur diesel fuel and are certified to EPA 2021 emissions standards. Key specifications of the engine are summarized in Table~\ref{tab:engine-specs}. The remaining sample engines share the same base architecture and emissions control configuration but exhibit engine-to-engine variability due to differences such as manufacturing imperfections and mechanical wear.

For one of the engines, referred to as the ``nominal engine'', both training and validation datasets were available. The training data were carefully designed to span the full range of operating conditions, ensuring the developed model is robust across all feasible engine states. This nominal engine corresponds to the development engine in the test program with the most extensive instrumentation and complete coverage of the operating space. The remaining three engines, referred to as ``sample engines'', are distinct production units of the same B6.7L platform, tested with the same fueling strategy, aftertreatment configuration, and fuel type. For the remaining three engines, referred to as ``sample engines'', only validation datasets were provided. 
These validation datasets were collected under conditions that mirrored the NOx levels and duty cycles of the validation data from the nominal engine.
Throughout this investigation, the experimental measurements serve as ground truth and provide the benchmark for model performance evaluation.

\begin{table}[htbp]
\centering
\caption{Specifications of the engine}
\label{tab:engine-specs}
\resizebox{8.5 cm}{!}{
\begin{tabular}{|c|c|} 
\hline
\textbf{Specification}            & \textbf{Details}                         \\ \hline
Engine type              & Cummins B6.7L CI engine                       \\
\hline
Horsepower                  & 200-325 hp (149--242 kW)                \\
\hline
Peak torque                & 520-750 lb-ft (705-1017 Nm)           \\
\hline
Governed speed                    & 2600 rpm                                 \\
\hline
Clutch engagement torque          & 400 lb-ft (542 Nm)                      \\
\hline
Number of cylinders               & 6                                        \\
\hline
Engine weight (dry)               & 1150 lb (522 kg)                        \\
\hline
Fuel system                       & High pressure common rail         \\
\hline
Turbocharger                      & VGT     \\
\hline
Emissions control                 & High pressure EGR \\
\hline
Certification                     & EPA 2021                                 \\ \hline
\end{tabular}}
\end{table}

\subsection{Data Normalization}

Accurately predicting instantaneous engine-out NOx presents notable challenges, primarily due to the continuous nature of NOx measurements and the prevalence of transient and extreme events. Conventional outlier removal methods, including box plot techniques or median based approaches as explored in \citep{yu2021novel, donateo2020real}, often exclude peak NOx emission events, which are critical for accurately modeling transient and rare operating scenarios.

Furthermore, standard normalization techniques such as min-max or standard scaler can inadequately handle extreme values, as these linear transformations remain significantly influenced by outliers. Consequently, we used the quantile transform normalization method \citep{peterson2020ordered}, which maps data onto a uniform or Gaussian distribution, thus providing robustness against outliers. By converting data points based on their relative ranking rather than their raw magnitude, quantile normalization effectively mitigates the disproportionate influence of extreme values, making it particularly suitable for complex, non-Gaussian emission datasets. All datasets used in this study were normalized using the quantile transform prior to GP model training on nominal engine data.

Gradient-based MCMC techniques, such as the No-U-Turn Sampler (NUTS) \citep{hoffman2014no}, are incompatible with the quantile normalization approach, primarily due to its nonlinear and discontinuous nature, which obstructs the gradient calculations essential for these methods. Although linear normalization could circumvent this challenge, it compromises predictive accuracy by inadequately addressing outliers present in the data. Thus, we use a gradient free ABC method, ideally suited for handling complex, non-differentiable components within our modeling framework.

\subsection{Metrics}

Model performance was evaluated using several statistical metrics carefully chosen to capture diverse aspects of prediction accuracy especially under extreme and regulatory critical conditions:

\begin{itemize}
\item \textbf{Root Mean Squared Error (RMSE)}: The primary metric used defined as
\[
\text{RMSE} = \sqrt{\frac{1}{N}\sum_{i=1}^{N}(y_i - \hat{y}_i)^2},
\]
effectively penalizes large prediction errors, which is essential for accurately capturing peak emission events.

\item \textbf{Percentiles of Absolute Errors (90th, 95th, 98th)}: To evaluate performance during challenging conditions, we computed these higher order percentiles of the absolute prediction errors:
\[
\text{AE}_{p} = \text{Percentile}_{p}(\{|y_i - \hat{y}_i|\}_{i=1}^{N}), \quad p \in \{90, 95, 98\},
\]
providing essential information on model reliability in infrequent but impactful emission scenarios.
\item \textbf{Coverage Probability of Credible Intervals:}
To assess uncertainty calibration, we computed the empirical coverage probability for the 95\% posterior predictive credible intervals (CI):
\[
\text{Coverage}_{95} = \frac{1}{N} \sum_{i=1}^{N} \mathbb{I}\left( y_i \in \left[\hat{y}_i^{2.5\%}, \hat{y}_i^{97.5\%}\right] \right),
\]
where $\hat{y}_i^{2.5\%}$ and $\hat{y}_i^{97.5\%}$ represents the lower and upper bounds of the interval, respectively. Well calibrated uncertainty estimates should yield coverage close to the nominal 95\%.

\end{itemize}

Additionally, to qualitatively assess model performance, we used cumulative NOx plotted over time. This visualization highlights the overall alignment between predicted and observed engine-out NOx, allowing intuitive identification of periods where the model consistently under or over estimates emissions.

\section{Results}
\label{sec:results}

The GP model was implemented using GPyTorch~\citep{gardner2018gpytorch} with PyTorch as the backend.  The loss function was defined as the negative of the exact marginal log-likelihood and the optimizer used is Adam~\citep{kingma2014adam} with tuned hyperparameters. To evaluate the ability of the models to accurately predict engine-out NOx, we used several quantitative metrics, including the RMSE and the 90th, 95th, and 98th percentiles of absolute errors. The parameters used in the proposed approach are detailed in Table \ref{tab:parameters_proposed_approach}.

\begin{table}[htbp]
\caption{Parameters used for proposed approach}
\label{tab:parameters_proposed_approach}
\resizebox{8.5 cm}{!}{%
\begin{tabular}{|l |r|}
\hline
\textbf{Distance metric}                   & Kolmogorov-Smirnov Statistic       
             \\ \hline
\textbf{Pilot sample size ($N_\text{pilot})$}                   & 1000                                                    \\ \hline
\textbf{Desired accepted samples ($N_\text{desired}$)}                   & 500                                                            \\ \hline
\textbf{Main sample size ($N_\text{main}$)}                   & 10000                                                           \\ \hline
\textbf{Data for inferring/estimating the biases}             & 200 s (FTP), 450 s (SET)    
            \\ \hline         
\textbf{Percentile $\zeta$}                   & 0.05                      
            \\ \hline
\textbf{Observation noise variance ($\sigma_y^2$)}                  & 0.01                              \\ \hline
\end{tabular}}
\end{table}

\begin{figure}[htbp]
    \centering
    \begin{subfigure}[b]{0.9\linewidth}
        \centering
        \includegraphics[width=\linewidth]{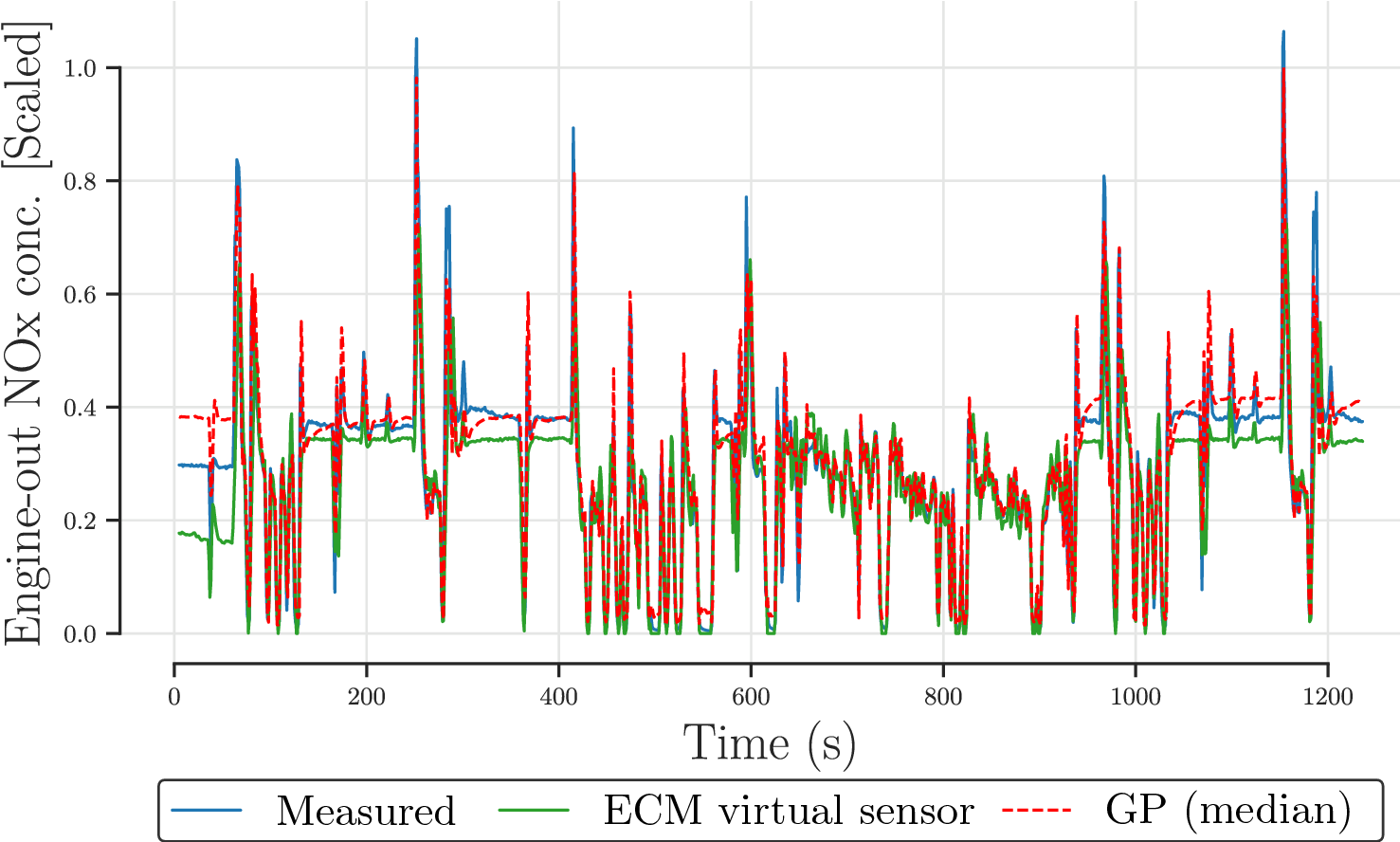}
        \caption{FTP cycle}
        \label{fig:val_1_ideal_engine}
    \end{subfigure}
    
    \vspace{0.5cm} 

    \begin{subfigure}[b]{0.9\linewidth}
        \centering
        \includegraphics[width=\linewidth]{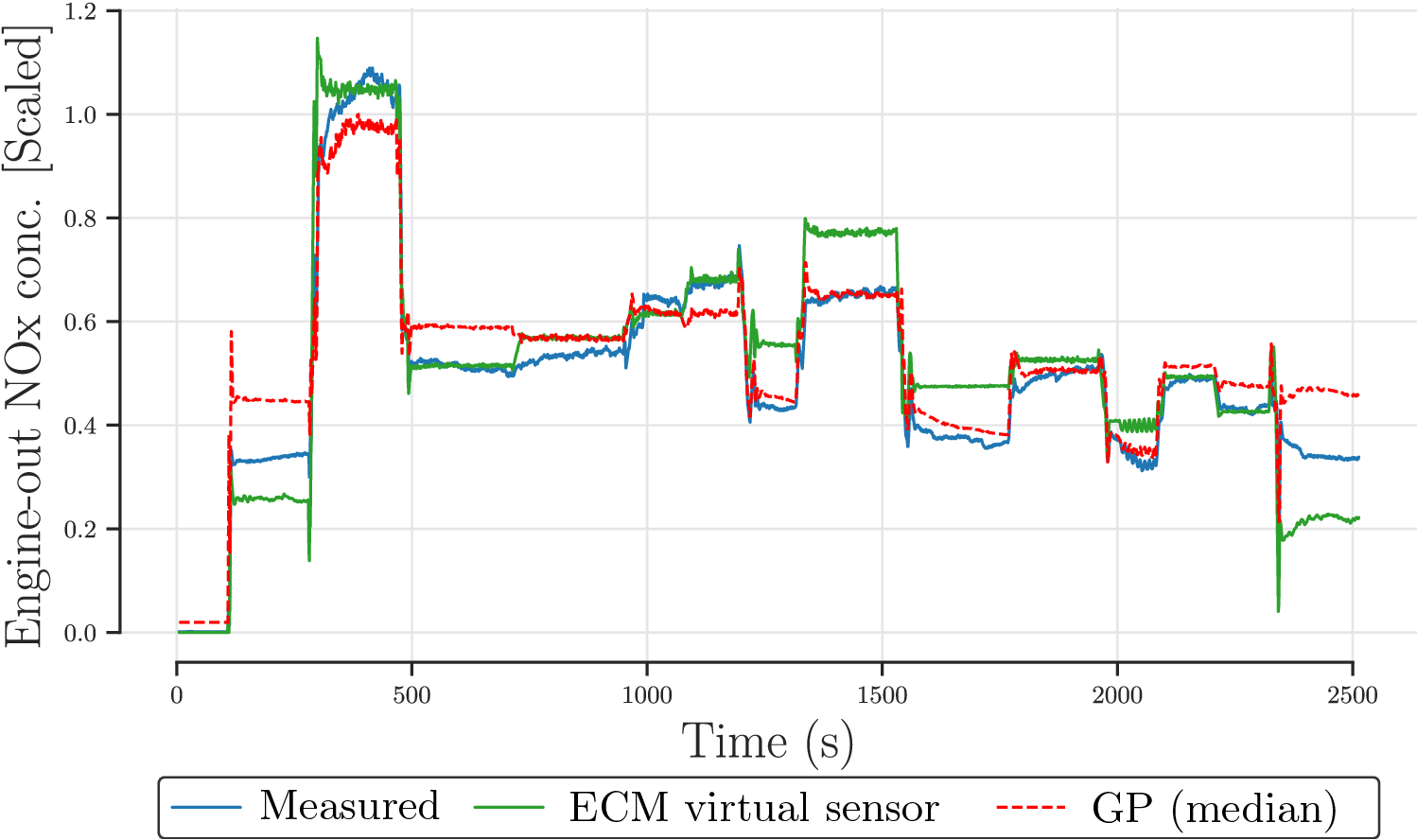}
        \caption{SET cycle}
        \label{fig:val_2_ideal_engine}
    \end{subfigure}

    \caption{GP predictions on validation datasets from nominal engine}
    \label{fig:gp_predictions_ideal_engine}
\end{figure}

All ABC experiments were executed on a desktop equipped with an Intel(R) Core(TM) i9-10900K CPU @ 3.70GHz and an NVIDIA RTX A4000 GPU with 16 GB of GDDR6 memory. Since the pretrained GP model remains fixed during calibration, the dominant computational cost arises from repeated GP median evaluations within the ABC simulation loop.

For each engine and cycle, the pilot phase used $N_{\text{pilot}} = 1000$ samples and required approximately 20 seconds of wall-clock time. The main phase used $N_{\text{main}} = 10000$ prior draws and required approximately 3–4 minutes per engine-cycle pair. The observed acceptance rate in the main phase was approximately $5\%$, consistent with the chosen quantile $\zeta = 0.05$.

In total, the ABC procedure required approximately $N_{\text{pilot}} + N_{\text{main}} = 11000$ parameter proposals per engine-cycle pair. Each proposal requires $T$ GP evaluations, resulting in approximately $2.2 \times 10^{6}$ GP forward evaluations for the FTP cycle ($T=200$) and $4.95 \times 10^{6}$ evaluations for the SET cycle ($T=450$). These evaluations are computationally efficient since they involve forward passes of a fixed GP model without gradient computation.

 The observation noise variance $\sigma_y^2 = 0.01$ was selected based on the empirical variance of the normalized NOx residuals from the pretrained GP model on nominal validation data. Sensitivity analysis with $\sigma_y^2 \in \{0.005, 0.01, 0.02\}$ resulted in negligible changes in posterior medians and predictive RMSE (less than 1\%), confirming robustness to moderate misspecification. A sensitivity study was also conducted by varying $N_{\text{pilot}} \in \{500, 1000, 2000\}$ and $N_{\text{main}} \in \{5000, 10000, 20000\}$. We observed that reducing $N_{\text{pilot}}$ below 1000 resulted in slightly noisier tolerance estimates but negligible change in posterior medians. Increasing $N_{\text{main}}$ beyond 10000 produced minimal improvement in predictive metrics (less than 1\% change in RMSE), indicating that the reported configuration in Table  \ref{tab:parameters_proposed_approach} achieves a stable tradeoff between computational cost and posterior accuracy.

Figure \ref{fig:gp_predictions_ideal_engine} illustrates the GP predictions on validation datasets derived from the nominal engine. We can see that the GP model, trained on experimental data from the nominal engine, outperformed the readings from a ECM virtual sensor provided by Cummins for comparison. 

\begin{figure}[htbp]
    \centering
    \includegraphics[width=0.99\linewidth]{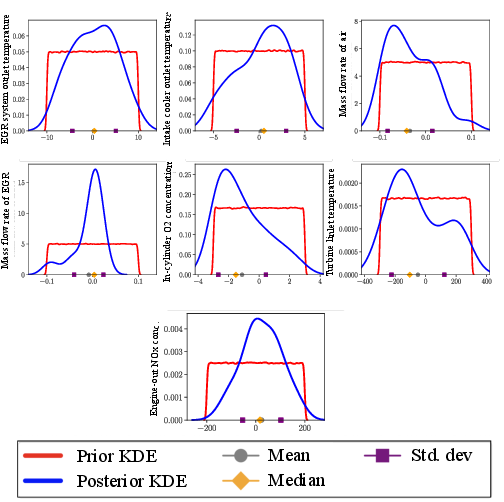}
    \caption{Inferred sensor biases for engine 1}
    \label{fig:biases_engine1}
\end{figure}

\begin{figure}[htbp]
    \centering
    \begin{subfigure}[b]{0.9\linewidth}
        \centering
        \includegraphics[width=\linewidth]{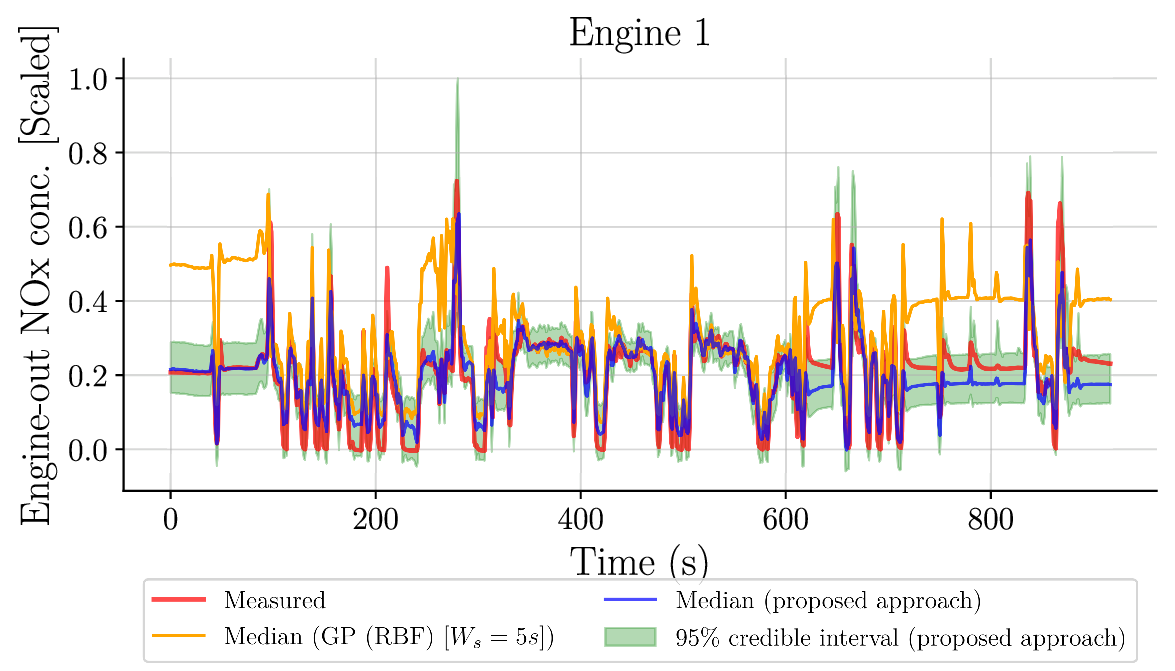}
        \caption{FTP cycle}
        \label{fig:predictions_engine1_1}
    \end{subfigure}%
    \hfill
    \begin{subfigure}[b]{0.9\linewidth}
        \centering
        \includegraphics[width=\linewidth]{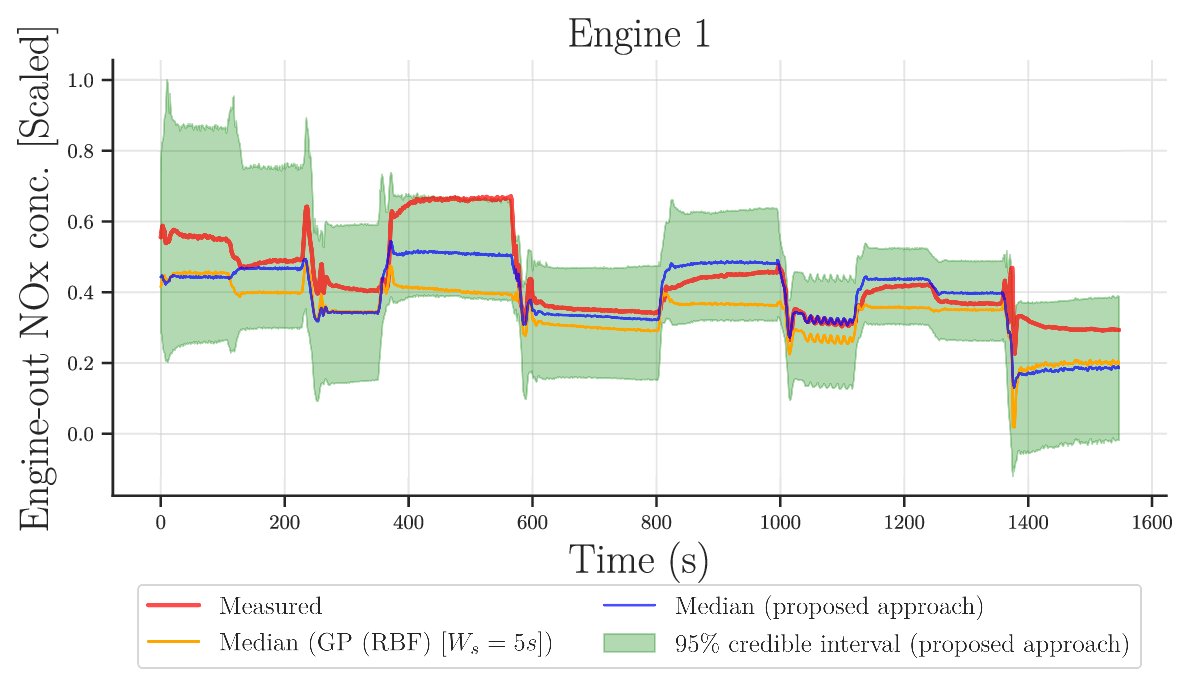}
        \caption{SET cycle}
        \label{fig:predictions_engine1_2}
    \end{subfigure}
    \caption{PPD of corrected model on unseen test data for engine 1}
    \label{fig:predictions_engine1}
\end{figure}

\begin{figure}[htbp]
    \centering
    \includegraphics[width=0.99\linewidth]{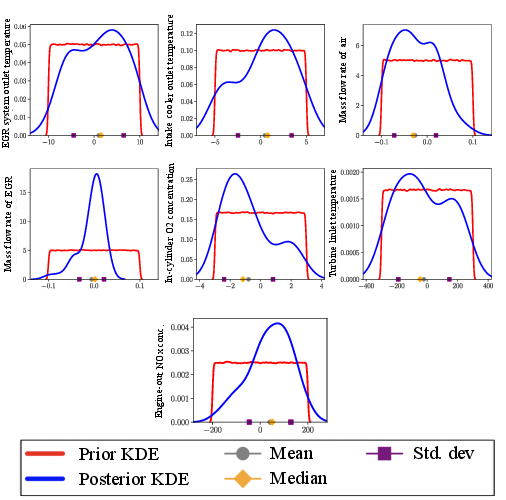}
    \caption{Inferred sensor biases for engine 2}
    \label{fig:biases_engine2}
\end{figure}

\begin{figure}[htbp]
    \centering
    \begin{subfigure}[b]{0.9\linewidth}
        \centering
        \includegraphics[width=\linewidth]{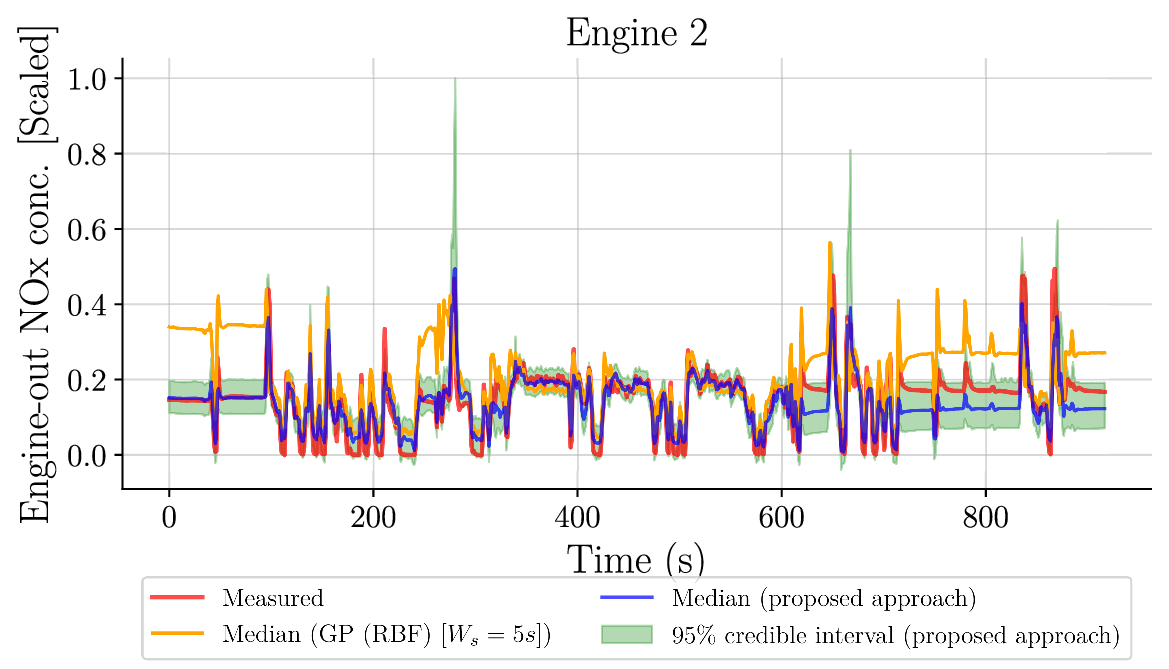}
        \caption{FTP cycle}
        \label{fig:predictions_engine2_1}
    \end{subfigure}%
    \hfill
    \begin{subfigure}[b]{0.9\linewidth}
        \centering
        \includegraphics[width=\linewidth]{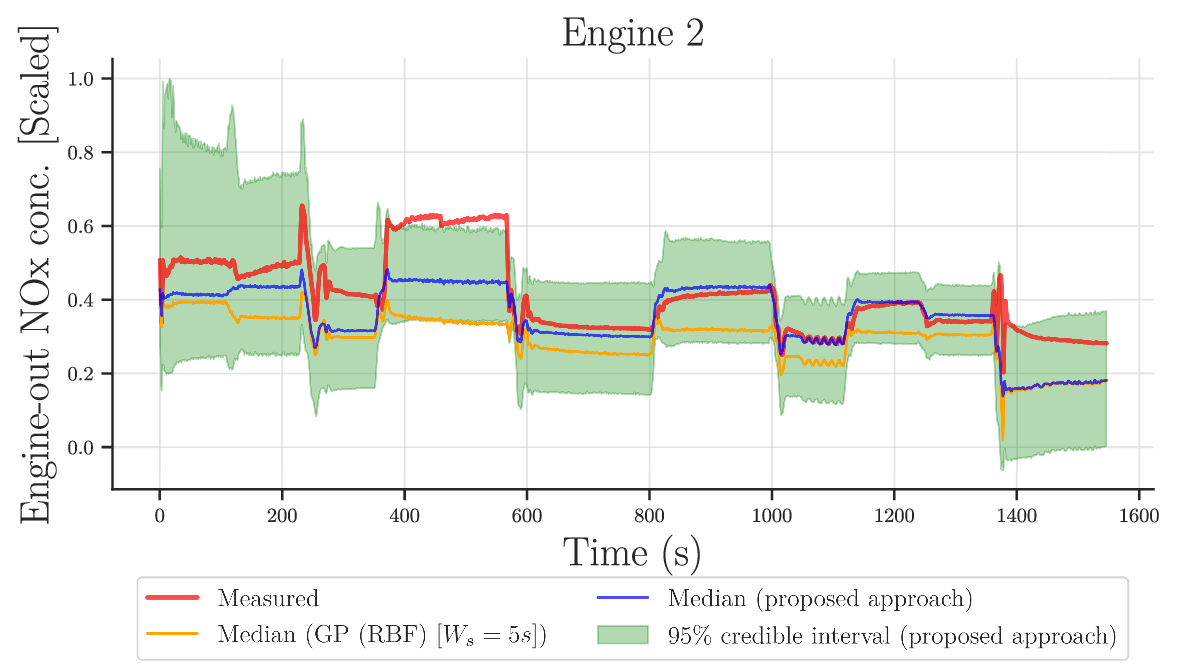}
        \caption{SET cycle}
        \label{fig:predictions_engine2_2}
    \end{subfigure}
     \caption{PPD of corrected model on unseen test data for engine 2}
    \label{fig:predictions_engine2}
\end{figure}

\begin{figure}[htbp]
    \centering
    \includegraphics[width=0.99\linewidth]{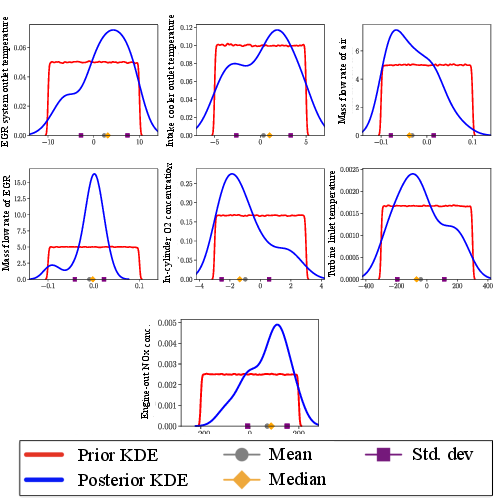}
    \caption{Inferred sensor biases for engine 3}
    \label{fig:biases_engine3}
\end{figure}

\begin{figure}[htbp]
    \centering
    \begin{subfigure}[b]{0.9\linewidth}
        \centering
        \includegraphics[width=\linewidth]{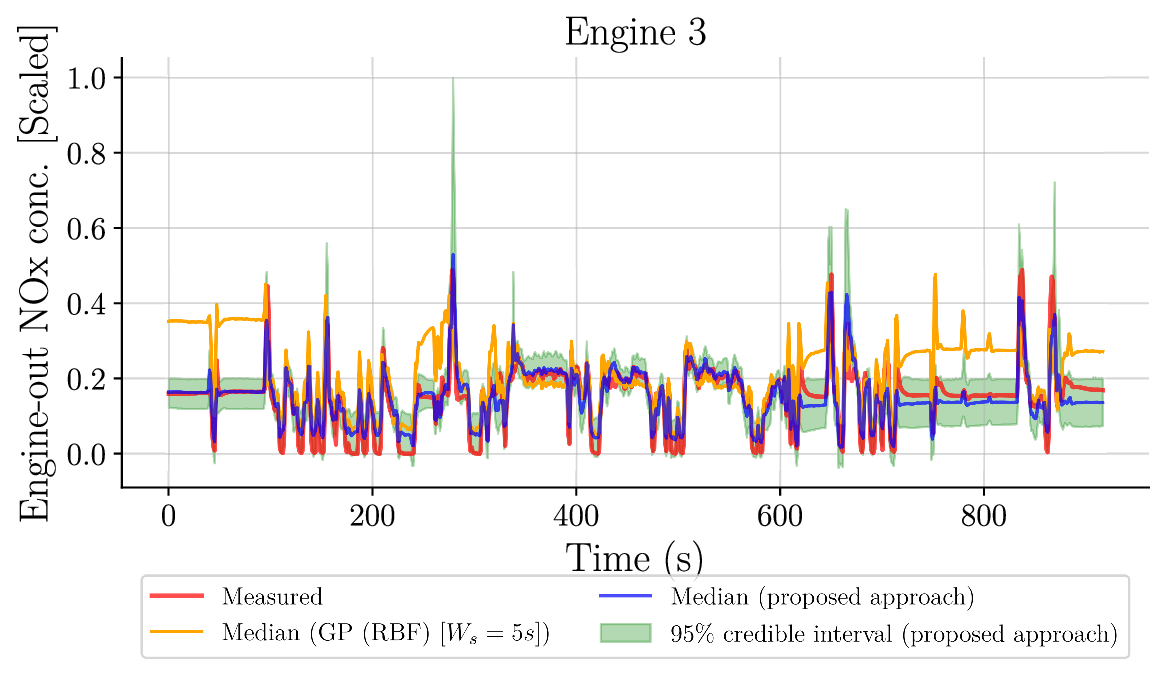}
        \caption{FTP cycle}
        \label{fig:predictions_engine3_1}
    \end{subfigure}%
    \hfill
    \begin{subfigure}[b]{0.9\linewidth}
        \centering
        \includegraphics[width=\linewidth]{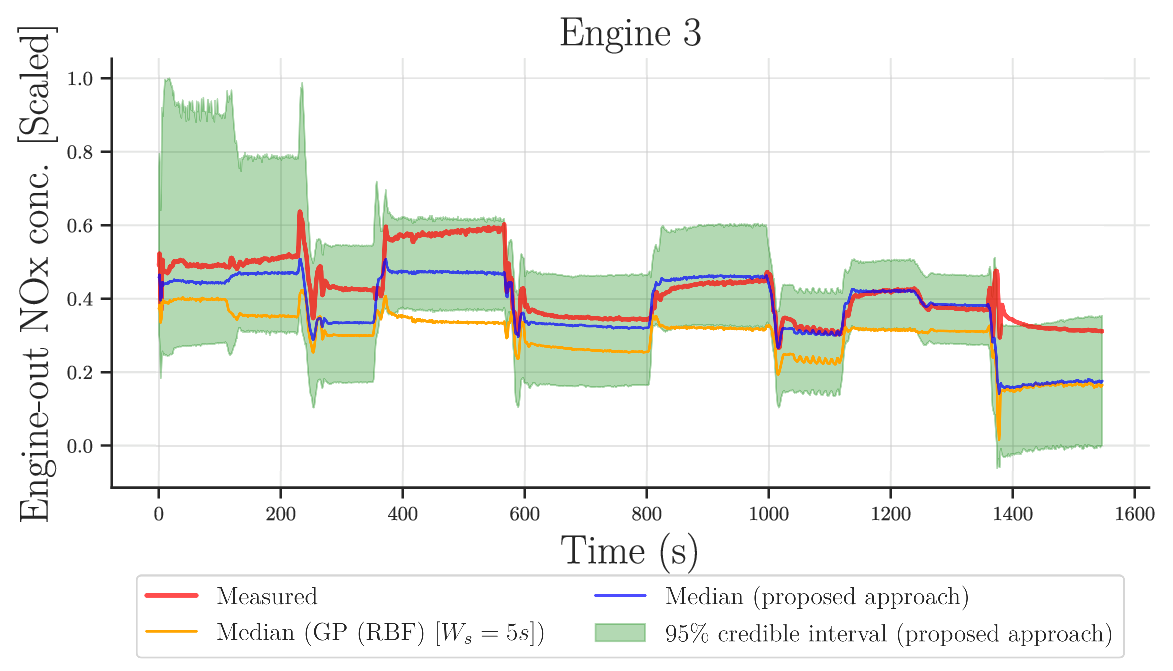}
        \caption{SET cycle}
        \label{fig:predictions_engine3_2}
    \end{subfigure}
    \caption{PPD of corrected model on unseen test data for engine 3}
    \label{fig:predictions_engine3}
\end{figure}

Note that since a larger error was observed in NOx concentrations during the initial 80 seconds of the FTP cycle and the first 400 seconds of the SET cycle, possibly caused due to one engine warming up faster than the other, this data was excluded from analysis, and the subsequent 200 seconds (FTP) and 450 seconds (SET) (as shown in Fig. \ref{fig:nox_different_engines}) were used to infer/estimate sensor biases. 
To quantify the impact of excluding warm-up data, we performed an ablation study where biases were inferred using the dataset including the initial 80 seconds of FTP and 400 seconds of SET versus using only the stabilized portion. Table \ref{tab:warmup_ablation} presents the comparison. We can see that including warm-up data degraded prediction accuracy by 14–25\% across all engines and cycles. This analysis revealed that during warm-up, several sensors exhibited non-stationary behavior inconsistent with the constant bias assumption, leading to biased parameter estimates that compromised predictions in the stabilized regime. These results justify our decision to exclude warm-up data for sensor bias calibration. It is important to note that the objective is not to infer/estimate the sensor biases correctly but to find sensor biases that improve the predictions of engine-out NOx.

\begin{table}[htbp]
\centering
\caption{Impact of including warm-up data on prediction accuracy}
\label{tab:warmup_ablation}
\resizebox{8.5 cm}{!}{
\begin{tabular}{|c|c|c|}
\hline
\multirow{2}{*}{\textbf{Calibration Data}} 
& \multicolumn{2}{c|}{\textbf{RMSE}} \\
\cline{2-3}
& \textbf{FTP cycle} & \textbf{SET cycle} \\
\hline
\multicolumn{3}{|c|}{\textbf{Engine 1}} \\
\hline
With warm-up (full) & 94.3 & 128.4 \\
\hline
Without warm-up (stabilized) & \textbf{70.5} & \textbf{110.3} \\
\hline
Improvement (\%) & 25.2\% & 14.1\% \\
\hline
\multicolumn{3}{|c|}{\textbf{Engine 2}} \\
\hline
With warm-up (full) & 108.7 & 135.2 \\
\hline
Without warm-up (stabilized) & \textbf{83.2} & \textbf{111.7} \\
\hline
Improvement (\%) & 23.5\% & 17.4\% \\
\hline
\multicolumn{3}{|c|}{\textbf{Engine 3}} \\
\hline
With warm-up (full) & 90.4 & 120.1 \\
\hline
Without warm-up (stabilized) & \textbf{68.8} & \textbf{100.2} \\
\hline
Improvement (\%) & 23.9\% & 16.6\% \\
\hline
\end{tabular}
}
\end{table}

Once the sensor biases were inferred/estimated using the proposed approach, these biases were used to perform posterior predictive checks on the remaining data. Figures 
\ref{fig:posterior_predictive_train_ftp} and \ref{fig:posterior_predictive_train_rmcset} show the PPD for the data used to infer sensor biases. This dataset, comprising 200 seconds from the FTP cycle and 450 seconds from the SET cycle for a specific sample engine, was used to select acceptable sensor biases based on the distance metric (i.e., KS statistic). We can see that the mean of the PPD for sample engines (engine 1, engine 2, and engine 3) closely matches the experimental data for the FTP cycle. However, for the SET cycle, the posterior mean exhibited relatively lower accuracy compared to the FTP cycle. This discrepancy aligns with the pretrained GP model’s reduced accuracy in predicting NOx concentrations for the SET cycle from the nominal engine, as shown in Fig. \ref{fig:val_2_ideal_engine}. Thus, such a pattern was expected for SET cycles from sample engines. The $95 \%$ credible intervals, presented in Figures \ref{fig:posterior_predictive_train_ftp} and \ref{fig:posterior_predictive_train_rmcset}, reflect the uncertainty arising from the stochastic nature of inferred biases. The SET cycle displayed greater uncertainty, indicating a higher sensitivity of the biases to steady-state conditions. This behavior is consistent with the lower baseline accuracy of the pre-trained GP on steady-state data (Fig. \ref{fig:val_2_ideal_engine}). Since the Bayesian calibration framework relies on the underlying model structure, the reduced fidelity in steady-state regimes naturally propagates into wider credible intervals in the PPD, ensuring that the model uncertainty accurately reflects the limitations of the base predictor.

\begin{table}[htbp]
    \centering
    \caption{FTP cycle}
    \label{tab:ftp_table}
    \resizebox{8.5 cm}{!}{
    \begin{tabular}{|c|c|c|c|c|}
    \hline
         \multirow{2}{*}{\textbf{Models}} & \multirow{2}{*}{\textbf{RMSE}} & \multicolumn{3}{c|}{\textbf{NOx Error Percentiles}} \\
         \cline{3-5}
         & & \textbf{90th} & \textbf{95th} & \textbf{98th} \\
         \hline
         \multicolumn{5}{|c|}{\textbf{Engine 1}} \\
         \hline
         GP (RBF) [$W_s = 5s$] & 209.76 & 390.08 & 399.92 & 441.88 \\
         \hline
         Proposed approach & \textbf{70.50} & \textbf{102.54} & \textbf{129.53} & \textbf{177.05} \\
         \hline
          \multicolumn{5}{|c|}{\textbf{Engine 2}} \\
         \hline
         GP (RBF) [$W_s = 5s$] & 197.68 & 388.41 & 392.33 & 435.44 \\
         \hline
         Proposed approach & \textbf{83.21} & \textbf{121.83} & \textbf{154.39} & \textbf{199.40} \\
         \hline
          \multicolumn{5}{|c|}{\textbf{Engine 3}} \\
         \hline
         GP (RBF) [$W_s = 5s$] & 193.27 & 371.43 & 378.69 & 393.61 \\
         \hline
         Proposed approach & \textbf{68.80} & \textbf{104.90} & \textbf{126.18} & \textbf{173.57} \\
         \hline
    \end{tabular}}
\end{table}

\begin{table}[htbp]
    \centering
    \caption{SET cycle}
    \label{tab:rmcset_table}
    \resizebox{8.5 cm}{!}{
    \begin{tabular}{|c|c|c|c|c|}
    \hline
         \multirow{2}{*}{\textbf{Models}} & \multirow{2}{*}{\textbf{RMSE}} & \multicolumn{3}{c|}{\textbf{NOx Error Percentiles}} \\
         \cline{3-5}
         & & \textbf{90th} & \textbf{95th} & \textbf{98th} \\
         \hline
         \multicolumn{5}{|c|}{\textbf{Engine 1}} \\
         \hline
         GP (RBF) [$W_s = 5s$] & 143.24 & 275.39 & 281.84 & 300.29 \\
         \hline
         Proposed approach & \textbf{110.34} & \textbf{207.16} & \textbf{215.25} & \textbf{229.22} \\
         \hline
          \multicolumn{5}{|c|}{\textbf{Engine 2}} \\
         \hline
         GP (RBF) [$W_s = 5s$] & 142.12 & 279.52 & 294.82 & 302.60 \\
         \hline
         Proposed approach & \textbf{111.65} & \textbf{208.23} & \textbf{217.72} & \textbf{226.15} \\
         \hline
          \multicolumn{5}{|c|}{\textbf{Engine 3}} \\
         \hline
         GP (RBF) [$W_s = 5s$] & 145.72 & 237.82 & 258.35 & 265.12 \\
         \hline
         Proposed approach & \textbf{100.17} & \textbf{181.05} & \textbf{193.54} & \textbf{227.76} \\
         \hline
    \end{tabular}}
\end{table}

\begin{figure*}[htbp]
    \centering
    \includegraphics[width = 16.8cm]{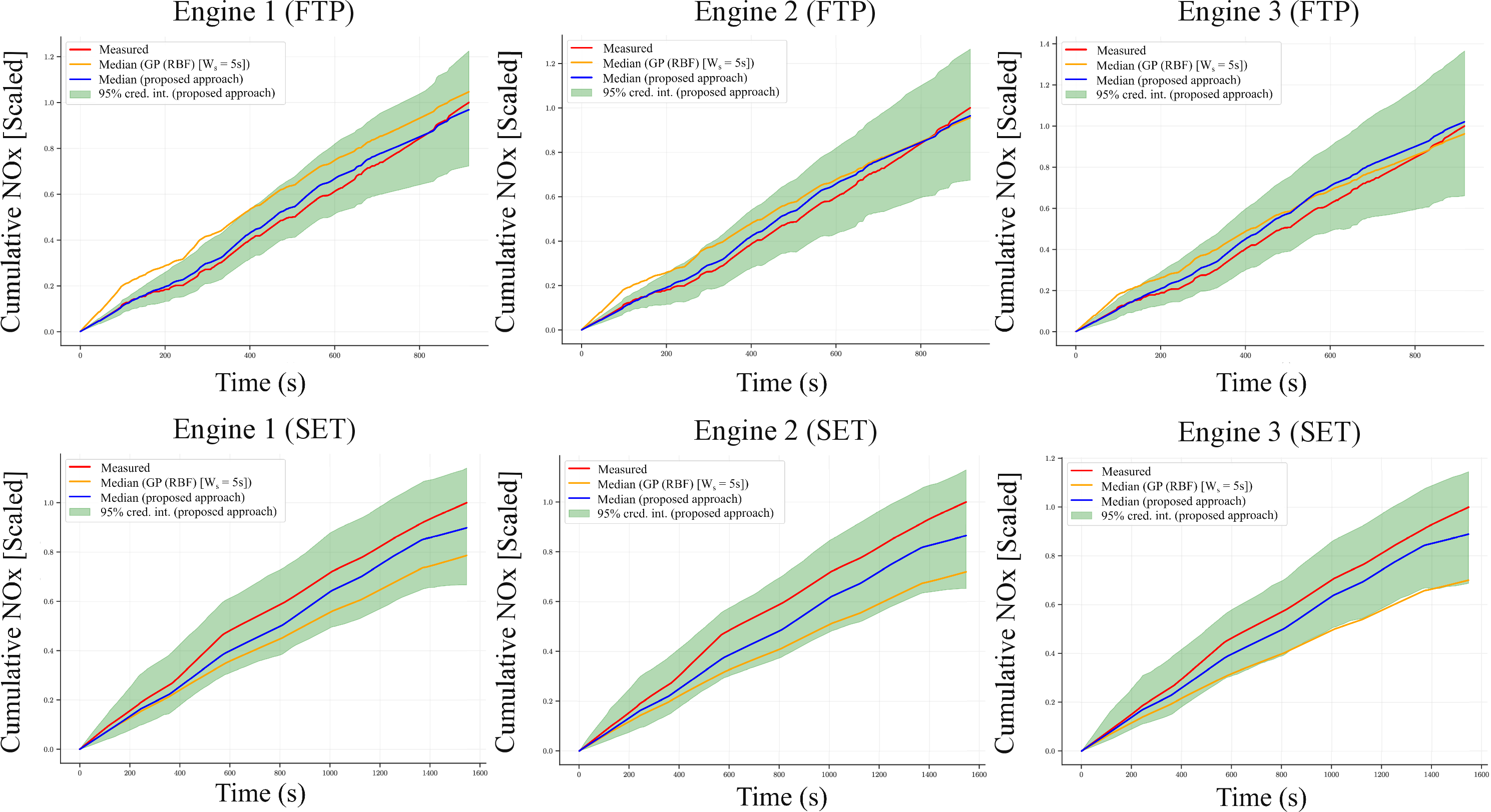}
    \caption{Cumulative engine-out NOx vs time}
     \label{fig:cum_nox_vs_time}
\end{figure*}

Fig. \ref{fig:biases_engine1} presents the kernel density estimations (KDEs) of the prior and posterior samples of the sensor biases for engine-out NOx and non-control input variables for engine 1. We can see that the posterior concentrates around the values, which minimizes the distance between the output of the simulated model and the data samples that were used to infer these biases. Fig. \ref{fig:predictions_engine1} compares the predictions of the proposed approach with the conventional pre-trained GP model (i.e, GP (RBF) [$W_s = 5s$]) for the remaining data across both cycles. The median predictions of the proposed approach more closely align with the experimental data than those from the pretrained GP model which was trained on nominal engine data. Where discrepancies existed, the measurements largely fell within the uncertainty bands. Similar observations can be made for engines 2 and 3, as illustrated in Figures \ref{fig:biases_engine2}, \ref{fig:predictions_engine2}, \ref{fig:biases_engine3}, and \ref{fig:predictions_engine3}.

Fig.~\ref{fig:cum_nox_vs_time} presents the cumulative engine-out NOx over time, comparing predictions from the proposed approach (corrected model) and the baseline GP model (GP (RBF) [$W_s = 5\,\mathrm{s}$]) against experimental measurements. Credible intervals representing uncertainty coming from engine-to-engine variability are also presented. We can see that the median predictions from the calibrated model consistently align more closely with measured data at all time steps compared to the baseline model highlighting improved predictive accuracy. Furthermore, the 95\% credible intervals of the proposed approach consistently includes the observed cumulative NOx measurements, emphasizing the robustness and reliability of the corrected probabilistic model in capturing uncertainty associated with engine specific variations.

To further validate the better performance of the proposed approach, Tables \ref{tab:ftp_table} and \ref{tab:rmcset_table} present the RMSE and NOx absolute error percentiles for both models. The results demonstrate significant improvements in these metrics for the proposed approach compared to the GP (RBF) [$W_s = 5s$].

\begin{table}[htbp]
\centering
\caption{Empirical coverage probability of 95\% posterior predictive credible intervals}
\label{tab:coverage_table}
\resizebox{8.5 cm}{!}{
\begin{tabular}{|c|c|c|}
\hline
\textbf{Engine} & \textbf{FTP Cycle (\%)} & \textbf{SET Cycle (\%)} \\
\hline
Engine 1 & 96.2 & 87.1 \\
\hline
Engine 2 & 96.1 & 86.3 \\
\hline
Engine 3 & 87.5 & 88.6 \\
\hline
\end{tabular}}
\end{table}

Table~\ref{tab:coverage_table} reports the empirical coverage probabilities of the 95\% posterior predictive intervals. For the FTP cycle, coverage is close to the nominal 95\% level across all engines, indicating well calibrated uncertainty. For the SET cycle, coverage remains above 85\%, with slight undercoverage relative to nominal levels. This behavior is consistent with the reduced fidelity of the pretrained GP in steady state dominated regimes and confirms that the proposed framework provides reliable, though not overly conservative, uncertainty quantification.

Unlike the baseline pretrained GP model, which suffers from significant performance degradation due to engine-to-engine variability, the proposed framework offers a robust solution with minimal data requirements. Specifically, by using only a short segment of engine-specific data (approximately 200 to 450 seconds) for calibration, our method achieves a drastic reduction in predictive error, lowering the RMSE by up to 63\% in transient cycles and 25\% in steady state cycles compared to the non adaptive GP model. This capability not only ensures high fidelity predictions across different engines but also allow rapid, scalable deployment by avoiding the computational burden of retraining models for individual units in a fleet.

\section{Conclusions}
\label{sec:conclusions}

This study introduced a Bayesian calibration framework aimed at improving the transferability of engine-out NOx predictive models across different engines. The calibrated predictions obtained through this methodology demonstrated substantial accuracy improvements compared to the baseline GP model, consistently observed across multiple engines and diverse operating cycles. Specifically, posterior predictive medians exhibited closer alignment with measured engine-out NOx, and their associated credible intervals effectively captured the uncertainty due to engine-to-engine variability. The cumulative NOx analysis and reduction in percentile errors further substantiated the robustness and practical value of the proposed approach for real world applications demanding cross engine generalizability.

Despite the significant improvements offered by our calibration method, several limitations must be considered:
\begin{itemize}
\item The computational demands of the ABC method grow significantly with increasing dimensionality, potentially limiting its scalability to scenarios involving extensive sensor configurations.
\item The framework relies on the accuracy of the underlying GP model. Reduced fidelity in specific operating regions, such as steady-state cycles, propagates into the calibrated predictions.
\item The current model assumes that the sensor and output biases ($\alpha_i, b_i$) are time invariant (constant offsets) over the entire operating cycle. While empirical evidence in Figs.~\ref{fig:inputs_cycle} and \ref{fig:nox_different_engines} supports this approximation over the analyzed windows, real-world deployment over extended durations may introduce time-varying bias due to sensor aging, thermal cycling, fouling, or actuator wear. In such cases, a static calibration may gradually lose accuracy as drift accumulates. The primary sources of error would arise from unmodeled slow bias drift rather than GP model error. Depending on drift magnitude, this could manifest as gradual increases in RMSE and systematic deviations in cumulative NOx over long operating horizons.
\end{itemize}

Potential extensions of this framework could address this by incorporating time dependent bias terms, potentially modeled via hierarchical GP or by implementing an online calibration strategy that periodically updates the bias estimates $(\alpha_i, b_i)$ using a sliding window of recent operating data.

Note that although this study focuses on engine out NOx, the proposed framework is more broadly applicable to systems in which a pretrained surrogate is deployed across a heterogeneous population and suffers from sensor specific biases. Examples include predictive models for other emission species such as particulate matter, carbon monoxide, or unburned hydrocarbons, as well as models for combustion phasing, turbocharger speed, or aftertreatment state estimation.

\paragraph{\normalfont{\textbf{Authors' contributions}}}
\hfill \\
\textbf{Shrenik Zinage:} Methodology, Software, Validation, Visualization, Writing - original draft. \textbf{Peter Meckl:} Funding acquisition, Writing - review and editing. \textbf{Ilias Bilionis:} Funding acquisition, Methodology, Writing - review and editing. 

\paragraph{\normalfont{\textbf{Acknowledgement}}}
\hfill \\
The authors thank Akash Desai of Cummins Inc. for his valuable feedback and guidance. They also acknowledge Dr. Lisa Farrell and Clay Arnett from Cummins Inc. for sponsoring this work, providing technical expertise, and providing critical experimental data for the simulations.

\paragraph{\normalfont{\textbf{Declaration of conflicting interests}}}
\hfill \\
The author(s) declared no potential conflicts of interest with respect to the research, authorship, and/or publication of this article.

\paragraph{\normalfont{\textbf{Declaration of generative AI and AI-assisted technologies}}}
\hfill \\
Portions of the text in this manuscript were refined using generative AI tools to improve clarity and readability. Specifically, ChatGPT was used to rewrite and polish sections of prose that were originally drafted by the authors, without introducing new conceptual content. In addition, Cursor was used as an AI-assisted coding environment to support code completion while generating the results reported in this study. All methodological designs, analytical decisions, interpretations, and conclusions were made solely by the authors. After using these tools, the authors reviewed and edited all AI-assisted outputs as needed and take full responsibility for the content of the publication.

\paragraph{\normalfont{\textbf{Funding}}}
\hfill \\
This work has been funded by Cummins Inc under grant number 00099056.

\bibliography{refs}

\section{Appendix}

\subsection{Approximate Bayesian Computation (ABC)}
\label{app:abc}

ABC~\citep{pritchard1999population} provides a framework for Bayesian inference when the likelihood is analytically intractable, yet generating synthetic data from the underlying model remains feasible. Formally, let us consider a probabilistic model $M$ characterized by parameters $\gamma \in \Theta$, with $\Theta$ representing the parameter space. Given observed data $D_\text{obs}$ generated from an unknown true process, standard Bayesian inference expresses the posterior distribution of parameters via the likelihood $p(D_\text{obs} | \gamma)$ and prior $p(\gamma)$ as:
\begin{equation*}
p(\gamma | D_\text{obs}) \propto p(D_\text{obs} | \gamma) p(\gamma).
\end{equation*}
In cases where direct evaluation of $p(D_\text{obs}|\gamma)$ is unfeasible, ABC avoids this limitation by relying exclusively on simulated data.

ABC operates by drawing parameter samples from the prior distribution $\gamma \sim p(\gamma)$ and subsequently simulating datasets $D_\text{sim}$ from the model $M$, according to the likelihood function $p(D|\gamma)$. To assess similarity between observed and simulated data, a predefined distance metric $\Delta(D_\text{sim}, D_\text{obs})$ is computed. Parameter samples are accepted if this distance metric does not exceed a predetermined threshold $\epsilon_\text{ABC}$; otherwise, they are rejected. Iterative application of this process yields an approximate representation of the posterior distribution.

If $\mathbb{I}[\cdot]$ represents the indicator function, the ABC posterior approximation is expressed as:
\begin{equation*}
\scalebox{0.85}{$
p_{\text{ABC}}(\gamma \mid D_{\text{obs}}, \epsilon)
\propto
\int
\mathbb{I}\!\left[\Delta(D_{\text{sim}}, D_{\text{obs}}) \leq \epsilon_{\text{ABC}}\right]
\, p(D_{\text{sim}} \mid \gamma)\, p(\gamma)\, dD_{\text{sim}}.
$}
\end{equation*}
The fidelity of the ABC posterior relies on the selected tolerance threshold $\epsilon_\text{ABC}$ and the choice of the distance metric $\Delta$. As $\epsilon_\text{ABC}$ approaches zero, under the condition that the distance metric comprehensively encodes all pertinent data information, the ABC posterior $p_{\text{ABC}}(\gamma | D_\text{obs}, \epsilon_\text{ABC})$ theoretically converges to the exact posterior $p(\gamma | D_\text{obs})$, as rigorously established in \citep{barber2015rate}. Practically, however, selecting $\epsilon_\text{ABC}$ poses a critical tradeoff: a smaller tolerance improves posterior accuracy but dramatically increases computational cost due to lower acceptance rates, whereas a larger tolerance reduces computational complexity but diminishes the accuracy of the posterior approximation.

\subsection{Kolmogorov-Smirnov Statistic (KS Statistic)}
\label{app:ks_statistic}

The KS statistic is a widely used nonparametric metric designed to measure the difference between two empirical probability distributions. In this paper, the KS statistic serves as a robust criterion to assess the discrepancy between simulated and observed time series datasets. Unlike conventional error metrics that focus on individual point deviations, the KS statistic evaluates the overall distributional similarity between datasets, thereby capturing holistic statistical characteristics.

The KS statistic is based on the ECDF. Given a dataset $X = \{x_1, x_2, \dots, x_n\}$, the ECDF denoted by $\hat{F}_n(t)$, is defined as the proportion of data points in $X$ that are less than or equal to a value $t$:
\begin{equation*}
\hat{F}_n(t) = \frac{1}{n}\sum_{i=1}^{n}\mathbb{I}[x_i \leq t].
\end{equation*}

Considering two independent samples, the observed data $D_{\text{obs}} = \{y_1, y_2, \dots, y_n\}$ and simulated data $D_{\text{sim}} = \{\hat{y}_1, \hat{y}_2, \dots, \hat{y}_m\}$, we compute their respective ECDFs, $\hat{F}_{D_{\text{obs}}}(t)$ and $\hat{F}_{D_{\text{sim}}}(t)$. The two sample KS statistic, represented by $D_{n,m}$, quantifies the maximal absolute difference between these two ECDFs over all possible values of $t$:
\begin{equation*}
D_{n,m} = \sup_{t} \left|\hat{F}_{D_{\text{obs}}}(t) - \hat{F}_{D_{\text{sim}}}(t)\right|.
\end{equation*}
A primary advantage of using the KS statistic is its nonparametric property, as it requires no assumptions regarding the underlying data distributions, such as normality. The statistic sensitively detects differences in central measures (such as mean) and distributional shape features (such as variance), providing it a comprehensive assessment tool for goodness of fit.

Compared to standard pointwise metrics such as the normalized root mean square error (NRMSE), the KS statistic is notably more robust. Although NRMSE accurately quantifies error magnitude at individual points, it is overly sensitive to minor temporal shifts common in dynamic systems. The KS statistic avoids such temporal sensitivity by evaluating statistical consistency across datasets, thus offering a more reliable and holistic evaluation of the model's fidelity.

\end{document}